\PassOptionsToPackage{table,xcdraw,dvipsnames}{xcolor}
\documentclass[10pt,twocolumn,letterpaper]{article}

\usepackage[pagenumbers]{iccv} 
\usepackage[accsupp]{axessibility}  

\usepackage{amsmath,amsfonts,bm}

\def\eqref#1{equation~\ref{#1}}

\def\1{\bm{1}}

\def\vc{{\bm{c}}}

\def\mW{{\bm{W}}}

\DeclareMathAlphabet{\mathsfit}{\encodingdefault}{\sfdefault}{m}{sl}
\SetMathAlphabet{\mathsfit}{bold}{\encodingdefault}{\sfdefault}{bx}{n}
\newcommand{\tens}[1]{\bm{\mathsfit{#1}}}

\def\tI{{\tens{I}}}

\def\tP{{\tens{P}}}

\def\tS{{\tens{S}}}

\DeclareMathOperator*{\sort}{sort}

\newcommand{\miou}{mIoU\xspace}
\newcommand{\method}{\text{FLOSS}\xspace}
\definecolor{infernoYellow}{RGB}{253,233,167}

\newcommand{\ourrow}{\rowcolor{infernoYellow}}
\newcommand{\ourcolorbox}[1]{\colorbox{infernoYellow}{#1}}

\newcommand{\class}{\ensuremath{\mathcal{C}}}

\newcommand{\visenc}{\ensuremath{f}}
\newcommand{\texenc}{\ensuremath{g}}
\newcommand{\template}{\ensuremath{\mathcal{T}}}
\newcommand{\prompt}{\ensuremath{\mathcal{P}}}
\newcommand{\avgemb}{\ensuremath{\bar{\vc}}}
\newcommand{\expert}{\ensuremath{\mathbf{\mathcal{E}}}}
\newcommand{\estexpert}{\ensuremath{\hat{\mathbf{\mathcal{E}}}}}

\newcommand{\PASCALCOFIVENINE}{PASCAL CONTEXT 59\xspace}
\newcommand{\PASCALCOFIVENINEabbr}{PC59\xspace}

\newcommand{\PASCALVOCTWENTY}{PASCAL VOC 20\xspace}
\newcommand{\PASCALVOCTWENTYabbr}{VOC20\xspace}

\newcommand{\ADETWENTYK}{ADE20K\xspace}
\newcommand{\ADETWENTYKabbr}{ADE\xspace}

\newcommand{\CITYSCAPES}{Cityscapes\xspace}
\newcommand{\CITYSCAPESabbr}{CS\xspace}

\newcommand{\ACDC}{ACDC\xspace}

\newcommand{\COCOSTUFF}{COCO-Stuff\xspace}
\newcommand{\COCOSTUFFabbr}{Stuff\xspace}

\newcommand{\BDD}{BDD-100K\xspace}
\newcommand{\BDDabbr}{BDD\xspace}

\newcommand{\MAPILLARY}{MAPILLARY\xspace}
\newcommand{\MAPILLARYabbr}{MAP\xspace}

\newcommand{\CLIPDINOISER}{CLIP-DINOiser\xspace}
\newcommand{\NACLIP}{NACLIP\xspace}
\newcommand{\MASKCLIP}{MaskCLIP\xspace}

\usepackage[normalem]{ulem}

\definecolor{iccvblue}{rgb}{0.21,0.49,0.74}
\usepackage[pagebackref,breaklinks,colorlinks,allcolors=iccvblue]{hyperref}

\usepackage{multirow}
\usepackage{textcomp}  
\usepackage{scalerel}  
\usepackage{makecell}
\usepackage{mathabx}
\usepackage[linesnumbered,ruled,vlined]{algorithm2e}
\usepackage{algpseudocode}
\usepackage[normalem]{ulem}
\usepackage{relsize}

\title{
FLOSS: Free Lunch in Open-vocabulary Semantic Segmentation\\}

\author{Yasser Benigmim$^{1}$\thanks{Work done during an internship at Inria.}
 \quad Mohammad Fahes$^{1}$ \quad  Tuan-Hung Vu$^{1,2}$ \quad Andrei Bursuc$^{1,2}$ \quad  Raoul de Charette$^{1}$
{}\\
$^1$ Inria \quad \quad \quad $^2$ Valeo.ai
}

\begin{document}
\newcolumntype{H}{>{\setbox0=\hbox\bgroup}c<{\egroup}@{}}

\maketitle

\begin{abstract}

In this paper, we challenge the conventional practice in Open-Vocabulary Semantic Segmentation (OVSS) of using averaged class-wise text embeddings, which are typically obtained by encoding each class name with multiple templates (e.g., \texttt{a photo of <class>}, \texttt{a sketch of a <class>}).
We investigate the impact of templates for OVSS, and find that for each class, there exist single-template classifiers---which we refer to as class-experts--- that significantly outperform the conventional averaged classifier.
First, to identify these class-experts, we introduce a novel approach that estimates them without any labeled data or training. By leveraging the class-wise prediction entropy of single-template classifiers, we select those yielding the lowest entropy as the most reliable class-experts. Second, we combine the outputs of class-experts in a new fusion process. Our plug-and-play method, coined \emph{\method{}}, is orthogonal and complementary to existing OVSS methods, offering an improvement without the need for additional labels or training.
Extensive experiments show that \emph{\method{}} consistently enhances state-of-the-art OVSS models, generalizes well across datasets with different distribution shifts, and delivers substantial improvements in low-data scenarios where only a few unlabeled images are available.
Our code is available at \url{https://github.com/yasserben/FLOSS}.

\end{abstract}
\section{Introduction}
\label{sec:intro}

In the past decade, advances in deep learning and the growing amount of training data have enabled the challenging task of semantic segmentation, which involves assigning semantic labels to each pixel in an image. Initially, this was limited to predefined categories~\cite{chen2017deeplab,strudel2021segmenter,xie2021segformer,cheng2022masked}, but recent models start using breakthroughs in vision-language alignment~\cite{radford2021learning} to move toward open-vocabulary segmentation, where the semantic categories can be defined dynamically at runtime, rather than being fixed in advance.

\begin{figure}[t]
    \centering
    \includegraphics[width=1.0\linewidth]{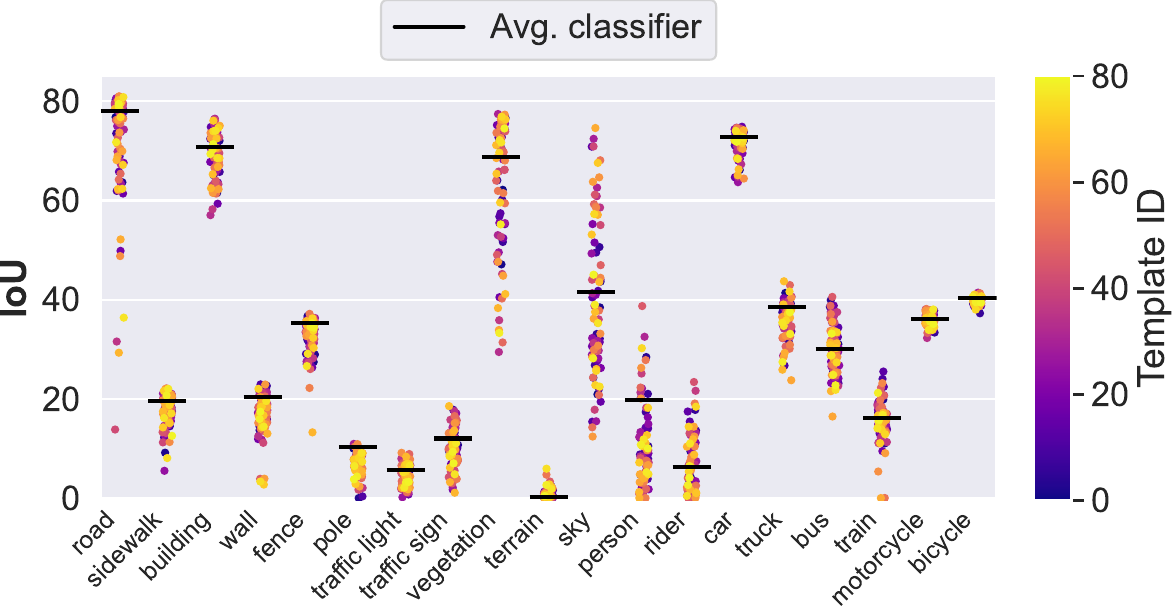}
    \caption{\textbf{Using individual vs average template classifiers.} Empirically, we observe that for each class there exist individual templates (colored dots) that lead to classifiers performing \textit{better} than the popular CLIP classifier, which is built from the averaging of all 80 templates embeddings (\textcolor{black}{\textbf{\textemdash}}). The analysis was conducted on~\CITYSCAPES using the \CLIPDINOISER model.}
    \label{fig:individualtemplates}
\end{figure}

In Open-Vocabulary Semantic Segmentation (OVSS), the CLIP model~\cite{radford2021learning} is commonly used as the backbone, which is either retrained~\cite{mukhoti2023open}, fine-tuned~\cite{liang2023open,cho2024cat}, or frozen and slightly modified to extend global image-language alignment to denser pixel-language alignment. In this work, we study CLIP's original classification paradigm and focus on methods that keep all CLIP parameters frozen~\cite{zhou2022extract,wang2024sclip,hajimiri2025pay}. Among these, MaskCLIP~\cite{zhou2022extract} leverages CLIP's embedding layer for segmentation, CLIP-DINOiser~\cite{wysoczanska2024clip} enhances features via distillation from an external DINO~\cite{caron2021emerging} network, while NACLIP~\cite{hajimiri2025pay} modifies CLIP's self-attention to improve spatial consistency.
A common aspect of prior OVSS methods is their reliance on CLIP's default classification process, where image or patch embeddings are compared with text embeddings representing semantic categories. To represent a class, a common practice is to use multiple templates to encode class names (\textit{e.g.}, ``a photo of a car'', ``a sketch of a car'', etc) and average their embeddings. Empirically, Radford \etal~\cite{radford2021learning} engineered $M=80$ templates to improve ImageNet zero-shot classification. For segmentation, OVSS methods~\cite{zhou2022extract,wysoczanska2024clip,hajimiri2025pay} adopt by default the same templates to construct segmentation classifiers. While this ``template-averaging''\footnote{``Template-averaging'' is an imprecise but convenient shorthand.} strategy works well overall, it may not always provide the best classifier for each semantic concept.

Our paper explores the impact of individual templates in OVSS -- an aspect that has been overlooked in the literature. \cref{fig:individualtemplates} shows an intriguing empirical observation: for each class, some individual templates outperform the averaged classifier that uses all templates. We call such templates \textbf{class-experts}, highlighting that they are trivial to identify using semantic ground truths.

However, without labels, two challenges arise: 1) \textit{How to identify 
class-experts without ground truth?} 2) \textit{How to fuse predictions from 
class-experts?}

We propose that entropy, an unsupervised metric, can help identify expert templates without labels. This leads to a new task---\textit{training- and label-free template selection for OVSS}---which improves segmentation by selecting better text template subsets for a given dataset of unlabeled images. Our method shifts the attention to the text modality, unlike previous works that tweak the visual encoder, and also aims to improve the performance in an inductive setting, where the selected templates are applied to a different subset of images.

\noindent We summarize our contributions as follows:
\begin{itemize}
    \item We analyze the current template-averaging practice for OVSS, showing that better class-wise classifiers can be constructed using only a subset of all templates, and that these expert templates can be identified without labels.
    \item We propose the novel task of training- and label-free template selection for OVSS. Given an unlabeled set of images, this task aims to identify class-wise experts that improve OVSS performance on unseen counterparts of the dataset.
    \item We introduce \method{}, which leverages class-wise per-pixel entropy to select expert templates and employs a simple and effective fusion scheme to combine individual expert predictions.

\end{itemize}
With extensive experiments, we demonstrate that \method{} consistently improves the state-of-the-art performance on OVSS benchmarks and exhibits generalization capabilities in the choice of experts.
Furthermore, with only a few unlabeled images, our approach can select reasonably good expert templates, resulting in performance boost.
\section{Related Works}
\label{sec:related}

\medskip\noindent\textbf{Vision-Language Model (VLM) transferability.}
Since CLIP's release, its impressive zero-shot classification performance~\cite{radford2021learning} and robustness~\cite{minderer2021revisiting,tu2023closer,tu2024empirical} have made it a foundation for various adaptation and transfer learning tasks. Few-shot supervised fine-tuning, where only limited labeled data is available for downstream tasks like image classification, has gained significant attention. The goal is to leverage the vast knowledge of the VLM acquired through training on an immense image collection by steering it towards a task of interest via few labeled samples.
Parameter efficiency is central to this setting. To ensure it, prompt learning~\cite{zhou2022learning,zhou2022conditional,chenplot,zhu2023prompt,khattak2023maple}, lightweight MLP training~\cite{gao2024clip}, and linear probing~\cite{huang2024lp++,silva2024closer} have been proposed. Another line of research explores training-free few-shot adaptation~\cite{zhang2022tip,zhu2023not,wang2024a}, where a classifier is built using few labeled samples and ensembled with the zero-shot classifier. Additionally, unsupervised adaptation has been explored~\cite{huang2022unsupervised,hu2024reclip,stojnic2024label}, alongside test-time adaptation, where prompts are learned per test image~\cite{shu2022test}.

While using VLMs for classification tasks aligns with their pre-training focus on global image recognition, applying them to dense prediction tasks like semantic segmentation is more challenging due to the gap between global image understanding and pixel-level prediction. We summarize related efforts in the following.

\medskip\noindent\textbf{Open-Vocabulary Semantic Segmentation (OVSS).}
OVSS uses VLMs like CLIP for text-based segmentation. However, CLIP's global pooling layers limit its ability to generate dense pixel-level, language-aligned features~\cite{jatavallabhula2023conceptfusion, zhou2022extract}. To address this, several \textit{training-free} methods densify CLIP without altering its parameters, preserving its image-language alignment. MaskCLIP~\cite{zhou2022extract} was the first to replace the attention pooling layer with a convolutional layer. Later works further densify CLIP by aggregating multiple image views~\cite{wysoczanska2024clip, jatavallabhula2023conceptfusion}, integrating priors from other vision foundation models for better pooling~\cite{wysoczanska2024clip, lan2024proxyclip}, extracting different information from the self-attention layers~\cite{bousselham2024grounding, li2025closer, hajimiri2025pay}, or removing anomaly tokens~\cite{bai2024self}.
Seg-TTO~\cite{de2025test} leverages an LLM to generate discriminative attributes for rare classes in specialized domains (\emph{e.g.}, medical sciences, earth monitoring) that use technical terminology (\emph{e.g.}, \emph{mediastinum}), operating in a test-time optimization framework while maintaining zero-shot capabilities.
Other approaches fine-tune CLIP for segmentation tasks using weak supervision through image-level tags, captions~\cite{ghiasi2022scaling,ranasinghe2023clippy}, or class-agnostic object masks~\cite{rao2021denseclip, ghiasi2022scaling}. Others employ full supervision~\cite{liang2023open,xu2023side,cho2024cat, li2022language} on densely annotated datasets like COCO Stuff~\cite{caesar2018coco}, following strategies similar to early inductive zero-shot semantic segmentation methods~\cite{xian2019semantic,bucher2019zero}. Among them, some fine-tune CLIP's image encoder~\cite{xu2023side,liang2023open,li2022language}, others combine this with additional backbones and segmentation heads~\cite{liang2023open}, while some learn side networks while keeping CLIP frozen~\cite{xu2023side}. Many other approaches have been proposed in this active research area~\cite{yu2023convolutions, xu2023open, xie2024sed, han2023open, xu2023learning}. While effective, these methods require labeled data and computational resources, limiting their practicality in resource-constrained scenarios.

Our work focuses on training-free approaches, evaluating FLOSS on three key methods: MaskCLIP~\cite{zhou2022extract}, CLIP-DINOiser~\cite{wysoczanska2024clip} and NACLIP~\cite{hajimiri2025pay}, while being applicable to any text-based segmentation framework.

\medskip\noindent\textbf{Prompting in VLMs.}
CLIP's image-language interface has inspired the adoption of various prompt engineering practices from LLMs to better reveal the knowledge encapsulated in the vision encoder and boost performance. The seminal CLIP work~\cite{radford2021learning} introduced dataset-specific templates (\textit{e.g.}, ``a photo of $<$class$>$'') to bridge the gap between training captions and test prompts for zero-shot classification. Subsequent methods have improved prompts by crowdsourcing templates~\cite{bach2022promptsource}, adjusting class names~\cite{ghiasi2022scaling, vobecky2023pop}, or modifying templates for tasks like object detection~\cite{gu2022open,ghiasi2022scaling} and occupancy prediction~\cite{vobecky2023pop}. While effective, manual prompt tuning is labor-intensive, leading to automated strategies like adding random words or characters to the original template~\cite{roth2023waffling}, enriching class names with WordNet concepts~\cite{ge2023improving} or synonyms~\cite{lin2023clip}, learning more informative class names~\cite{huang2024renovating}, and using LLMs to generate more expressive descriptions starting from simple class names or retrieve them from a dataset~\cite{rubin2022learning} and further use them as prompts. Some approaches also learn prompts for task-specific fitting~\cite{parisot2023learning} or test-time adaptation~\cite{shu2022test}, and others mine them with external LLMs~\cite{esfandiarpoor2024if}.

Our approach differs by selecting expert templates for each class in an unsupervised manner, starting from predefined templates. While previous methods have focused on image classification, we apply this strategy to semantic segmentation using CLIP's original ImageNet templates, which have been used by default in prior works.
\section{Preliminaries}
\label{sec:preliminaries}

OVSS aims
to produce a soft-segmentation map \mbox{$\hat{\tS}\in [0,1]^{H \times W \times K}$} given an image \mbox{$\tI \in \mathbb{R}^{H \times W \times 3}$} and a set of classes $\{\class_k\}_{k \in [1,K]}$ expressed in natural language.
The final segmentation map $\hat{\tP}$ is derived by applying $\arg\max(\cdot)$ on $\hat{\tS}$ along the $K$-dimension.
Leveraging the aligned vision encoder $\visenc{}(\cdot)$ and text encoder $\texenc{}(\cdot)$ of CLIP, the standard practice consists in classifying an image patch by comparing its vision embedding to the text embedding of class-specific prompts constructed using predefined templates $\{\template_m\}_{m \in [1,M]}$. An example of prompt is ``$\underbrace{\text{a bright photo of a}}_{\template_m}$ $\underbrace{\text{car}}_{\class_k}$'', denoted $\prompt_{m,k}$ for compactness. {Instead of a single template, to increase robustness and improve overall performance, practitioners average the embeddings of prompts obtained using all available templates~\cite{radford2021learning,roth2023waffling,menon2023visual,pratt2023does}, which is a form of ensembling on the text encoder side. This writes $\avgemb_k = \frac{1}{M} \sum\limits_{m=1}^{M} \texenc(\prompt_{m,k})$, where $\avgemb_k$ is the resulting representation of class $\class_k$.
Subsequently, for each patch, the predicted class is the one that maximizes the cosine similarity between the visual and the text embeddings:
\begin{equation}
    \hat{k} = \arg \max_{k} f(\mathcal{V})^{\mathrm{T}} \avgemb_k\,,
    \label{eq:cossim}
\end{equation}
where $\mathcal{V}$ represents the visual patch. All embeddings are $\ell_{2} \text{-normalized}$. The text embeddings of $K$ classes define a classifier that we denote:
\begin{equation}
    {\mW}(\{\mathcal{T}_m\}_{m \in [1, M]}) = \{\bar{\vc}_k\}_{k\in [1,K]}\,.
    \label{eq:default_classif}
\end{equation}

\noindent {As in the above equation, the classifier can be represented as a function of the templates used to compute the average embeddings of all classes.}

\medskip\noindent\textbf{Empirical observations on single templates.}
\label{sec:obs_expert}
{OVSS models~\cite{zhou2022extract,wysoczanska2024clip,hajimiri2025pay,xie2024sed} typically rely on the $M{=}80$ ImageNet templates of CLIP~\cite{radford2021learning} to compute the average classifier of~\cref{eq:default_classif}, which we refer to as ${\mW}_\text{CLIP}$ for brevity. While using ${\mW}_\text{CLIP}$ was shown to be robust~\cite{radford2021learning, zhou2022extract}, the templates were originally hand-crafted for zero-shot classification on ImageNet~\cite{radford2021learning}, and their individual effect on class-wise performance remains unexplored for OVSS.}
Rather than ${\mW}_\text{CLIP}$ that uses all $M$ templates,
one can
construct $M$
distinct classifiers,
each utilizing a single template $\mathcal{T}_m$. This writes:
\begin{equation}
    \mW(\mathcal{T}_m) = \{g(\prompt_{m,k})\}_{k \in [1,K]}
    \label{eq:single_classif}
\end{equation}

We design a systematic experiment to evaluate the performance of single-template classifiers $\mW(\mathcal{T}_m)$ from~\cref{eq:single_classif}.
\cref{fig:individualtemplates} shows
the per-class performance on \CITYSCAPES~\cite{cordts2016cityscapes} of each of the $80$ single-template classifiers
(plotted as colored dots, \eg,~\textcolor{orange}{\raisebox{-0.4ex}{\scalebox{1.5}{\textbullet}}}), where colors encode the template identifier~(\ie,~$m$),
along with the average ${\mW}_\text{CLIP}$ classifier~(plotted as~\textcolor{black}{\textbf{\textemdash}}).
This leads to two key observations:

\noindent\textbf{a)}~For each class, there exist single-template classifiers outperforming ${\mW}_\text{CLIP}$ \textit{on this class};
we refer {to} {the corresponding templates} as \mbox{\textbf{class-experts}}. More precisely, the set of class-experts is defined as:
\begin{equation}
    \expert_k = \{\mathcal{T}_m \mid \Omega_k(\mW(\mathcal{T}_m))>\Omega_k({\mW}_\text{CLIP}), m \in [1,M]\}\,,
    \label{eq:experts}
\end{equation}
where $\Omega_k(\cdot)$ denotes the IoU performance on class $k$.

\noindent \textbf{b)}~A single-template classifier excelling on one class may exhibit suboptimal performance on others, implying that for $k,l \in [1, K]$ the expert sets $\expert_{k}$ and $\expert_{l}$ \textit{may} differ.

Following the above observations, class-experts are readily identified when ground truth labels are accessible, through direct performance comparison.
We refer readers to~\cref{sec:empirical-obs} for similar observations across datasets and models.

\noindent However, we are interested in the case where an unlabeled dataset is provided. Therefore, two questions stem from the previous empirical observations:
(i)~\textit{How to identify class-experts in an unsupervised, training-free manner?} (ii)~\textit{Given these identified class-experts, how to derive the final semantic predictions?}
\begin{figure*}[t!]
    \centering
    \includegraphics[width=\linewidth, trim=14mm 0mm 1mm 0mm, clip]{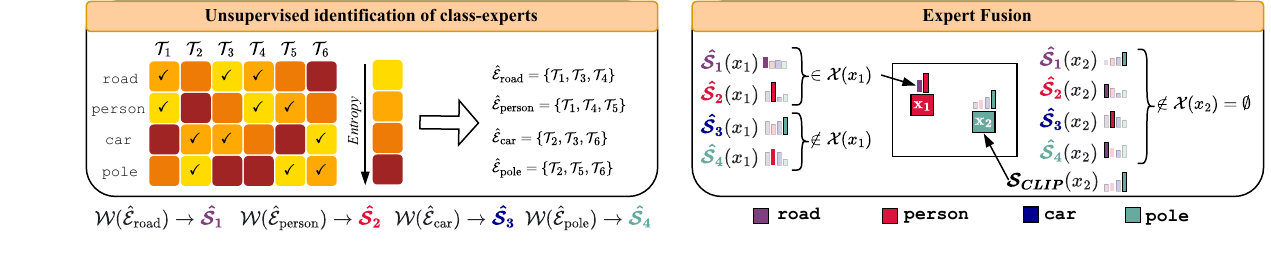}
    \caption{\textbf{Overview of \method{}.} \textbf{Left:} for each class, selected expert templates ($\checkmark$) construct the estimated set $\estexpert$, which classifies the input image and yield the soft-segmentation map $\tS$. In this example we set $N=3$ for the \texttt{Top-N(\dots)} template indices corresponding to lowest entropy entries to select. \textbf{Right:} we explain the fusion strategy for two cases. For the pixel $x_1$, the set of experts that predict their own class of expertise $\mathcal{X}(x_1)$ contains 2 experts; consequently, the decision is taken by looking at the maximum softmax scores of those two, leading to the decision of classifying $x_1$ as ``person'' with the higher score. For the pixel $x_2$, there is no expert predicting its own class of expertise, i.e., $\mathcal{X}(x_2) = \emptyset$, and the decision is taken by the default classifier ${\mW}_\text{CLIP}$; the pixel $x_2$ thus takes the scores from ${\tS}_\text{CLIP}(x_2)$.}
    \label{fig:approach}
\end{figure*}

\section{Method}
\label{sec:method}
Given access to a dataset $\mathcal{D}$ of unlabeled images, our goal is to improve the performance of OVSS using the aforementioned class-experts, \textit{without training or access to any labels}.
To do so, for each class $k \in [1, K]$ our method first identifies a small set of class-experts among the $M$ pre-defined ImageNet templates of CLIP~(\cref{sec:estimation}) by leveraging unsupervised metrics and then builds on a simple scheme to fuse class-experts' predictions into a single OVSS semantic output~(\cref{sec:fusion}).

\subsection{Unsupervised identification of class-experts}
\label{sec:estimation}
Referring to our definition of class-experts in~\cref{sec:obs_expert}, for each of the $K$ classes, our goal is to estimate a set of class-experts $\estexpert_k$, among the $M$ pre-defined CLIP templates.
Being in an unsupervised setting, we have no access to the true classifier performance (\ie, $\Omega(\cdot)$) and therefore propose relying on unsupervised metrics which act as performance proxy~\cite{wang2021tent, xie2024mano}.
We leverage entropy as it is an established measure of uncertainty~\cite{hullermeier2021aleatoric,gal2016uncertainty}, commonly adopted in deep learning literature~\cite{kendall2017uncertainties,lakshminarayanan2017simple,vu2019advent}.
The entropy value of a softmax prediction indicates a classifier's confidence, which empirically provides a reliable indicator of prediction quality. In our setting, entropy measures the confidence of single-template classifiers for each class, helping identify class-experts.
We also study other unsupervised metrics in the ablations (\cref{sec:ablation}).

In detail, we estimate class-experts as the set of individual templates whose associated classifiers (\cf, \cref{eq:single_classif}) exhibit low entropy. Specifically, for each template~$m$ and class~$k$, we compute the average entropy of the $C_{m,k}$ pixels predicted as class~$k$ by $\mW(\mathcal{T}_m)$ in all images of $\mathcal{D}$:
\begin{equation}
    \mathcal{H}_{m,k} = -\frac{1}{C_{m,k}}\sum_{i=1}^{C_{m,k}} \mathbf{q}_i^{\mathrm{T}} \, \text{log}(\mathbf{q}_i)\,,
    \label{eq:entropy}
\end{equation}
where $\mathbf{q}_i \in \Delta^{K-1}$ is pixel-wise probability prediction obtained by applying  $\texttt{softmax}(\cdot)$ over $K$ class-wise cosine similarities, and $\Delta^{K-1}$ is the $(K-1)$-simplex in $\mathbb{R}^K$.
Subsequently, for class $k$, the estimated set of experts is obtained as the $N$ templates yielding the lowest entropy:
\begin{align}
    \label{eq:estexpert}
    \estexpert_k = \{\mathcal{T}_{\hat{m}} \mid \hat{m} \in \texttt{Top-N}(\arg\sort_{m} \mathcal{H}_{m,k}) \}\,,
\end{align}

\noindent{}where $\estexpert_k$ is expected to be an estimation of $\expert_k$ (\cref{eq:experts});
the $\arg\underset{m}{\sort}$ operator returns the sorted list of $M$ template indices so that $\mathcal{H}_{m,k}$ are in ascending order. Finally, $\texttt{Top-N}(\cdot)$ thus returns N template indices corresponding to lowest entropy entries.
In practice, we use $N=4$ and validate this choice through an ablation study in our experimental section.
Following~\cref{eq:default_classif}, the expert classifier of class $k$, denoted ${\mW}(\estexpert_k)$, is constructed by averaging the embeddings obtained from the $N$ templates in $\estexpert_k$ for each of the $K$ class names.

The above process leads to $K$ classifiers
$\{{\mW}(\estexpert_k)\}_{k \in [1,K]}$, all obtained in a fully unsupervised manner, and each is expected to excel on one specific class.
However, it is important to note that although a classifier is expert of a single class, it outputs a full semantic map containing $K$ classes, and the above formulation produces $K$ soft-segmentation maps, $\{\hat{\tS}_k\}_{k\in[1,K]}$. To output a single OVSS map, we introduce a simple scheme to fuse all expert predictions.

\subsection{Fusion of expert predictions}
\label{sec:fusion}
To fuse all predicted soft-segmentation maps $\{\hat{\tS}_k\}_{k \in [1,K]}$ into a single output $\hat{\tS}$, we rely on the fact that each map $\hat{\tS}_k$ excels at segmenting class $k$.
Therefore, for each pixel, the class assigned is the one having the highest probability among the expert classifiers \textit{which predicted their own class of expertise}.
Occasionally, when no expert predicts its own class of expertise for a given pixel,
we simply revert to using the ${\mW}_\text{CLIP}$ classifier.
In practice, we observe that only \mbox{$\approx2\%$}~of pixels fall
under the latter case. Therefore, the vast majority of decisions are carried out by experts.

Formally, for a given pixel $x$, $\hat{\tS}_k(x)$ is the $K-$dimensional softmax probability of the expert $k$ at the position of $x$. Let \mbox{$\mathcal{X}(x) = \{k\mid\underset{i \in [1,K]}{\arg\max}~\hat{\tS}_k(x)[i] = k\}$} be the index set of experts that predicted their own class of expertise for $x$. The prediction of $x$ is determined as:
\begin{equation}
    \hat{\tP}(x) =
    \begin{cases}
        k^* = \underset{k \in \mathcal{X}(x)}{\arg\max}~\hat{\tS}_k(x)[k],    & \text{if } \mathcal{X}(x) \neq \emptyset, \\
        k^\dagger = \underset{k \in [1,K]}{\arg\max}~{\tS}_\text{CLIP}(x)[k], & \text{otherwise}.
    \end{cases}
\end{equation}
Fig.~\ref{fig:approach} illustrates the unsupervised identification of class-experts as well as the fusion strategy.
\noindent Overall, our method incurs computational overhead primarily from computing cosine similarities for the $K$~class-experts, which scales with the number of classes and becomes more significant when $K$ is very large ($>$100). However, this overhead is constrained to the similarity computations since the vision encoder is shared across all experts, requiring only a single forward pass regardless of the number of classes. We provide more details in ~\cref{sec:inf_gpu}.
\section{Experiments}
\label{sec:exps}

\paragraph{Experimental Setup.} {We evaluate our method on three state-of-the-art open-vocabulary semantic segmentation models:
    the seminal \MASKCLIP~\cite{zhou2022extract} and two recent  models \CLIPDINOISER~\cite{wysoczanska2024clip}
    and  \NACLIP~\cite{hajimiri2025pay}.
    \CLIPDINOISER trains a convolution layer using only 1K unlabeled images from ImageNet~\cite{deng2009imagenet} for fast inference on pixel-level segmentation tasks, effectively imitating the DINO priors for pooling guidance and removing the need for DINO encoder at runtime. \MASKCLIP and \NACLIP are training-free adaptation methods of CLIP.

    \noindent For a fair comparison, we report the version of \NACLIP without refinement. We primarily select training-free models that keep the image encoder frozen and produce unaltered, dense image-language features. This enables us to assess the impact of the class-experts identification strategy by using the original CLIP pre-trained features and ImageNet templates.
    Unless otherwise stated, all models utilize CLIP ViT-B/16 backbone. \CLIPDINOISER and \MASKCLIP are based on OpenCLIP~\cite{cherti2023reproducible}, while \NACLIP employs OpenAI's original CLIP~\cite{radford2021learning}}. For all models -- \CLIPDINOISER, \NACLIP, and \MASKCLIP-- we follow the authors' evaluation protocol. For \CLIPDINOISER and \MASKCLIP, we resize images to 448 pixels on the shortest side with a sliding window of size 448 and stride 224. For \NACLIP, images are resized to 336 pixels (except for \CITYSCAPES which are resized to 560 pixels) with a 224 window and stride 112.

\noindent\textbf{Datasets and Metric.} We evaluate FLOSS across multiple semantic segmentation benchmarks: \PASCALCOFIVENINE~\cite{mottaghi2014role}~(\PASCALCOFIVENINEabbr), \PASCALVOCTWENTY~\cite{pascal-voc-2012}~(\PASCALVOCTWENTYabbr), \COCOSTUFF~\cite{caesar2018coco}~(\COCOSTUFFabbr), \ADETWENTYK~\cite{zhou2019semantic}~(\ADETWENTYKabbr), and \CITYSCAPES~\cite{cordts2016cityscapes}~(\CITYSCAPESabbr). {Note that these datasets exhibit different ontology and semantic granularity, ranging from 19 classes (\CITYSCAPESabbr) to 171 classes (\COCOSTUFFabbr).}
Additionally, we demonstrate the generalization capabilities of our class-experts identified on \CITYSCAPES to other driving datasets {such as \BDD~\cite{yu2020bdd100k} and \MAPILLARY~\cite{neuhold2017mapillary} as well as \ACDC~\cite{sakaridis2021acdc} under Night, Fog, Rain and Snow conditions}.
{Importantly, our approach is training-free and uses no labels. Performance is measured using mean Intersection over Union (mIoU).}

{\smallskip\noindent\textbf{Implementation details.} The only hyperparameter of \method{}, being the number of class-experts to identify in~\cref{eq:estexpert}, is set to $N=4$ in all experiments.
While our method is training-free, class-experts are estimated using the training set of each dataset, while performance is always reported on the corresponding validation set.}

\subsection{Main results}
\begin{table}[t]
    \centering
    \resizebox{\linewidth}{!}{%
        \begin{tabular}{lccccc|c} %
            \toprule
            \textbf{Method}                          & \textbf{\CITYSCAPESabbr} & \textbf{\PASCALVOCTWENTYabbr} & \textbf{\PASCALCOFIVENINEabbr} & \textbf{\ADETWENTYKabbr} & \textbf{\COCOSTUFFabbr} & \textbf{Avg}  \\
            \midrule
            CLIP~\cite{radford2021learning}          & 6.7                      & 49.1                          & 11.2                           & 3.2                      & 5.7                     & 15.2          \\
            GroupViT~\cite{xu2022groupvit}           & 11.1                     & 79.7                          & 23.4                           & 9.2                      & 15.3                    & 27.7          \\
            CLIP-Surgery~\cite{li2025closer}         & 31.4                     & --                            & --                             & 12.9                     & 21.9                    & --            \\
            GEM~\cite{bousselham2024grounding}       & --                       & --                            & 32.6                           & 15.7                     & --                      & --            \\
            SCLIP~\cite{wang2024sclip}               & 32.2                     & 80.4                          & 34.2                           & 16.1                     & 22.4                    & 37.1          \\
            CLIP-DIY~\cite{wysoczanska2024clipdiy}   & 11.6                     & 79.7                          & 19.8                           & 9.9                      & 13.3                    & 26.9          \\
            TCL~\cite{cha2023learning}               & 23.1                     & 77.5                          & 30.3                           & 14.9                     & 19.6                    & 33.1          \\
            ReCo~\cite{shin2022reco}                 & 21.1                     & 57.8                          & 22.3                           & 11.2                     & 14.8                    & 25.4          \\
            \midrule
            \MASKCLIP~\cite{zhou2022extract}         & 25.0                     & \textbf{61.8}                 & 25.5                           & 14.2                     & 17.5                    & 28.7          \\
            \ourrow+ \method{}                       & \textbf{25.8}            & \textbf{61.8}                 & \textbf{26.2}                  & \textbf{14.9}            & \textbf{17.8}           & \textbf{29.3} \\
            \midrule
            \NACLIP~\cite{hajimiri2025pay}           & 35.5                     & 79.7                          & 35.2                           & 17.4                     & 23.3                    & 38.2          \\
            \ourrow+ \method{}                       & \textbf{37.0}            & \textbf{80.2}                 & \textbf{35.9}                  & \textbf{18.4}            & \textbf{23.6}           & \textbf{39.0} \\
            \midrule
            \CLIPDINOISER~\cite{wysoczanska2024clip} & 31.3                     & 80.9                          & 35.9                           & 20.0                     & 24.6                    & 38.5          \\
            \ourrow+ \method{}                       & \textbf{34.6}            & \textbf{82.3}                 & \textbf{36.2}                  & \textbf{20.7}            & \textbf{24.7}           & \textbf{39.7} \\
            \bottomrule
        \end{tabular}%
    }
    \caption{{\textbf{OVSS across datasets and models.} We report mIoU metric for eight OVSS baselines and three models either using the average templates (\ie, \MASKCLIP, \NACLIP, and \CLIPDINOISER) or with our method (\ie, \mbox{+\method{}}). \method{} consistently improves OVSS models} across datasets of varying complexity, from urban scenes (\CITYSCAPES, 19 classes) to general objects (\COCOSTUFF, 171 classes). Bold highlights the \textbf{best} performance.}
    \label{tab:sota_table}
\end{table}
\cref{tab:sota_table} reports the mIoU of OVSS baselines across five datasets showing \method{} plugged into three different models: \MASKCLIP, \NACLIP and \CLIPDINOISER.
The performance gains vary with dataset complexity, showing smaller improvements on \PASCALVOCTWENTYabbr $[+0.05, +0.5, +1.4]$ or \PASCALCOFIVENINEabbr $[+0.7, +0.7, +0.3]$ than on~\CITYSCAPESabbr $[+0.8, +1.5, +3.3]$. We attribute this to \PASCALVOCTWENTYabbr and \PASCALCOFIVENINEabbr being closer to ImageNet which was leveraged in CLIP~\cite{radford2021learning} to engineer the set of $80$ templates used in all models.
The benefit of our method is evident on challenging benchmarks like \ADETWENTYKabbr and \COCOSTUFFabbr, which have a large number of semantic categories making accurate segmentation particularly hard.

\smallskip\noindent\textbf{Quality of experts.} {We evaluate the quality $\hat{\rho}_k$ of our experts by measuring, for each class $k$, the intersection of the estimated class-expert $\estexpert{}_k$ with the true set of experts~$\expert_k$:}
\begin{equation}
    \hat{\rho}_k = 100\times\frac{|\estexpert{}_k \cap \expert{}_k|}{|\estexpert{}_k|} = 100\times\frac{|\estexpert{}_k \cap \expert{}_k|}{N}\,.
    \label{eq:experts_accuracy}
\end{equation}
Where $|.|$ denotes the cardinality of a set. Quality averaged over classes is reported in \cref{tab:accuracy} and shows that regardless of the model used, we correctly estimate approximately half of the true experts. In some scenarios, such as \PASCALVOCTWENTY with \NACLIP, even when only 20\% of predicted experts are true experts, it is sufficient to bring some improvements (+0.5 mIoU in \cref{tab:sota_table}).
Per-class quality is reported in~\cref{sec:supp-expert-accuracy}, showing some variability.
\begin{table}[t]
    \centering
    \newcommand{\var}[1]{}
    \resizebox{\linewidth}{!}{%
        \begin{tabular}{lccccc|c} %
            \toprule
            \textbf{Method}          & \textbf{\CITYSCAPESabbr} & \textbf{\PASCALVOCTWENTYabbr} & \textbf{\PASCALCOFIVENINEabbr} & \textbf{\ADETWENTYKabbr} & \textbf{\COCOSTUFFabbr} & \textbf{Avg} \\
            \midrule
            \MASKCLIP+ \method{}     & 51\%\var{39}             & 54\%\var{35}                  & 56\%\var{35}                   & 57\%\var{37}             & 56\%\var{35}            & 55\%         \\
            \NACLIP+ \method{}       & 49\%\var{35}             & 20\%\var{30}                  & 57\%\var{37}                   & 57\%\var{37}             & 52\%\var{39}            & 47\%         \\
            \CLIPDINOISER+ \method{} & 62\%\var{37}             & 41\%\var{36}                  & 57\%\var{32}                   & 45\%\var{37}             & 44\%\var{36}            & 50\%         \\
            \bottomrule
        \end{tabular}%
    }
    \caption{{\textbf{Quality of our estimated class-experts.} We measure the quality of our estimated experts by evaluating their intersection with the true experts, \cf~\cref{eq:experts_accuracy}.}}
    \label{tab:accuracy}
\end{table}

\smallskip\noindent\textbf{Generalization to unseen datasets.}
\begin{table}[t]
    \centering
    \setlength{\tabcolsep}{4pt}
    \resizebox{\linewidth}{!}{%
    \begin{tabular}{lc|cccccc|c}
    \toprule
    \multirow{2}{*}{\textbf{Method}} & in domain & \multicolumn{7}{c}{out of domain}\\
     & \textbf{\CITYSCAPESabbr} & \textbf{Night} & \textbf{Fog} & \textbf{Rain} & \textbf{Snow} & \textbf{\BDDabbr} & \textbf{\MAPILLARYabbr} & \textbf{Avg}\\
    \midrule
    MaskCLIP &  24.5 & 13.3 & 20.3 & 21.0 & 19.9 & 22.7 & 25.8 & 20.5\\
     \ourrow+ \method{} & \textbf{25.8} & \textbf{13.9} &  \textbf{21.7} & \textbf{22.3}  & \textbf{21.3} & \textbf{22.9}& \textbf{26.8} & \textbf{21.5} \\
    \midrule
    NACLIP & 35.5  & 23.3 & \textbf{32.9}  & 30.7 &32.6 & 31.4 & 35.3 & 31.0 \\
     \ourrow+ \method{} & \textbf{37.0} & \textbf{23.6} &  32.5  & \textbf{31.0} & \textbf{33.6} & \textbf{32.3} & \textbf{36.5} & \textbf{31.6} \\
    \midrule
    CLIP-DINOiser & 31.3  & 12.7  & 24.1  & 27.9 & 27.2 & 31.3 & 35.2 & 26.4 \\
     \ourrow+ \method{} & \textbf{34.6} & \textbf{16.6} & \textbf{29.0} & \textbf{29.8} & \textbf{29.4} & \textbf{32.2} & \textbf{36.6} & \textbf{28.9} \\
    \bottomrule
    \end{tabular}}
    \caption{{\textbf{Generalization of experts across datasets.} We evaluate the transferability of experts identified on the \CITYSCAPES (\CITYSCAPESabbr) dataset to unseen datasets that share the same semantic classes and exhibit changes in lighting (Night), weather (Fog, Rain, Snow), or location and scenery (\BDDabbr, \MAPILLARYabbr). Results show a consistent improvement when using \method{}, across almost all settings. }
    }
    \label{tab:tab_generalization}
\end{table}{We now evaluate the generalization of experts to other unseen datasets sharing the same label set but exhibiting distribution shifts. In \cref{tab:tab_generalization}, we report performance of \method{} when experts are identified on \CITYSCAPES and used to evaluate performance on six unseen datasets which include four ACDC conditions (Night, Fog, Rain, Snow), \BDD, and \MAPILLARY.}
{Overall, we consistently improve the performance with notable gains of up to $+4.9$ \miou for \CLIPDINOISER on ACDC Fog.}
{This demonstrates that our class-wise experts are readily usable on other datasets sharing the same semantic classes, despite the presence of distribution shifts.}

\smallskip\noindent\textbf{Low-data regimes.} To investigate the data efficiency of \method{}, we study its performance in lower data regimes. \cref{fig:num_images_ablation} reports the mIoU of \CLIPDINOISER\ on \CITYSCAPES and \PASCALVOCTWENTY,  as the number of images sampled from their respective training sets (containing 2,975 and 1,464 images, respectively) varies. The figure shows the average and standard deviation over 10 seeds (i.e., randomly sampling 10 different image sets).
On both datasets, the general trend shows that our method benefits from accessing more images, which stems from the better estimation of class-experts.
Remarkably, our method performs well in a very low-data regime, outperforming the default \CLIPDINOISER with only 1 image on \CITYSCAPES and around 25 images on \PASCALVOCTWENTY.
We conjecture that the latter requires more images due to its lower per-image class density compared to \CITYSCAPES, which exhibits higher variability and density per image.

\begin{figure}[t]
    \centering
    \begin{subfigure}[b]{0.49\columnwidth}
        \centering
        \includegraphics[width=\textwidth]{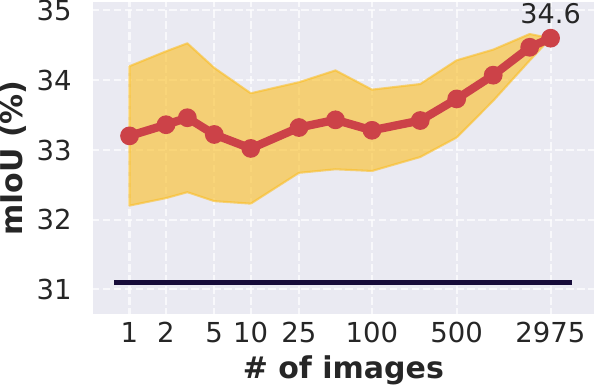}
        \caption{\CITYSCAPES}
        \label{fig:ablation_cityscapes}
    \end{subfigure}%
    \hfill%
    \begin{subfigure}[b]{0.49\columnwidth}
        \centering
        \includegraphics[width=\textwidth]{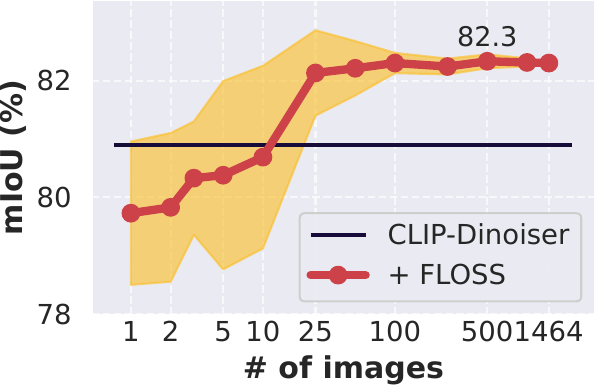}
        \caption{\PASCALVOCTWENTY}
        \label{fig:ablation_pascal}
    \end{subfigure}
    \caption{{\textbf{Effect of the number of images.} We evaluate \CLIPDINOISER with \method{}  on urban scenes~\subref{fig:ablation_cityscapes} and general object segmentation~\subref{fig:ablation_pascal}, when varying the number of images used to identify class-experts. Results show the average of 10 seeds, with standard deviation. Despite larger variation, we note the robustness of our method, even when having access to very few unlabeled images.}}
    \label{fig:num_images_ablation}
\end{figure}

\smallskip\noindent\textbf{Low-data regimes \& domain shift.} To further corroborate its practicality, we evaluate FLOSS under stricter conditions combining both low-data regimes and domain shift. Specifically, we use only a single CS image to identify experts that are tested on 6 shifted domains. \cref{tab:tab_one_image_and_shift} provides evidence on the effectiveness of FLOSS in this strict setting, leading to an average improvement of 1.9\% across OOD datasets.

\smallskip\noindent\textbf{Expert transferability.} We investigate the cross-dataset transferability of class-experts by evaluating whether experts identified on one dataset can effectively improve performance on semantically related but distinct datasets. We employ \COCOSTUFF (171 classes) to identify class-experts for \CLIPDINOISER and apply them to \CITYSCAPES, \PASCALVOCTWENTY, and \PASCALCOFIVENINE, which share 10, 16, and 37 common classes with \COCOSTUFF, respectively. For non-overlapping classes, we use the default average classifier ${\mW}_\text{CLIP}$. As reported in~\cref{tab:tab_unrelated_images}, FLOSS shows consistent improvements, confirming expert transferability across datasets.

\begin{table}[t]
    \centering
    \setlength{\tabcolsep}{4pt}
    \resizebox{\linewidth}{!}{%
        \begin{tabular}{lc|cccccc|c}
            \toprule
            \multirow{2}{*}{\textbf{Method}} & in domain                & \multicolumn{7}{c}{out of domain}                                                                                                               \\
                                             & \textbf{\CITYSCAPESabbr} & \textbf{Night}                    & \textbf{Fog}  & \textbf{Rain} & \textbf{Snow} & \textbf{\BDDabbr} & \textbf{\MAPILLARYabbr} & \textbf{Avg}  \\
            \midrule
            CLIP-DINOiser                    & 31.3                     & 12.7                              & 24.1          & 27.9          & 27.2          & 31.3              & 35.2                    & 26.4          \\
            \ourrow+ \method{}               & \textbf{33.2}            & \textbf{16.2}                     & \textbf{28.2} & \textbf{29.3} & \textbf{28.7} & \textbf{31.7}     & \textbf{35.6}           & \textbf{28.3} \\
            \bottomrule
        \end{tabular}}
    \caption{{\textbf{FLOSS with 1 image \& domain shift.} Experts are identified using \textbf{a single \CITYSCAPES (\CITYSCAPESabbr) image} and evaluated on unseen datasets that share the same semantic classes and exhibit changes in lighting (Night), weather (Fog, Rain, Snow), or location and scenery (\BDDabbr, \MAPILLARYabbr).
            }}
    \label{tab:tab_one_image_and_shift}
\end{table}
\begin{table}[t]
    \centering
    \setlength{\tabcolsep}{4pt}
    \resizebox{\linewidth}{!}{%
        \begin{tabular}{lc|cc|cc|cc}
            \toprule
            \multirow{3}{*}{\textbf{Method}} & \textbf{in-domain}      & \multicolumn{6}{c}{\textbf{out of domain}}                                                                                                                                                            \\
                                             & \textbf{\COCOSTUFFabbr} & \multicolumn{2}{c}{\textbf{\CITYSCAPESabbr}} & \multicolumn{2}{c}{\textbf{\PASCALVOCTWENTYabbr}} & \multicolumn{2}{c}{\textbf{\PASCALCOFIVENINEabbr}}                                                 \\

                                             &                         & \textbf{com.}                                & \textbf{all}                                      & \textbf{com.}                                      & \textbf{all}  & \textbf{com.} & \textbf{all}  \\
            \midrule
            CLIP-DINOiser                    & 24.6                    & 34.5                                         & 31.1                                              & 80.5                                               & 80.8          & 39.3          & 36.0          \\
            \ourrow + \method{}              & \textbf{24.7}           & \textbf{35.8}                                & \textbf{32.2}                                     & \textbf{82.3}                                      & \textbf{82.4} & \textbf{40.0} & \textbf{36.2} \\
            \bottomrule
        \end{tabular}}
    \caption{{\textbf{Cross-dataset expert transferability.} Class-experts identified on COCO-Stuff are applied to CS, VOC20, and PC59. Results are reported for common classes shared with COCO-Stuff (\textbf{com.}) and all classes (\textbf{all}). FLOSS demonstrates consistent improvements across all target datasets.}}
    \label{tab:tab_unrelated_images}
\end{table}

\subsection{Ablation Study}
\label{sec:ablation}

\paragraph{Fusion strategies for expert predictions.}
{We investigate alternative strategies to fuse the expert predictions (\cf,~\cref{sec:fusion}), reporting performance on \CITYSCAPES in \cref{tab:fusion_strategies}.}
Simply averaging all soft segmentation maps (``Average-all'') irrespective of the {templates class of expertise} leads to suboptimal results across all models.
When accounting for {templates' expertise in fusion, conflicts arise often as \textit{several} experts may predict their own class of expertise. We therefore explore three strategies to resolve conflicts:} ``Default'' which falls back to ${\mW}_\text{CLIP}$ predictions, ``Average'' which averages the probability vectors of conflicting experts, and ``Highest'' which selects the most confident expert as described in~\cref{sec:fusion}. The latter consistently yields the best performance across all models, validating our fusion scheme.

\smallskip\noindent\textbf{Metrics for expert identification.}
{We replace entropy in~\cref{eq:estexpert} with other unsupervised metrics to act as proxies of the template class-wise performance, reporting performance on~\CITYSCAPES using the \CLIPDINOISER model in~\cref{tab:model_comparison}.}

\noindent(i) \emph{Avg. Probability} computes the average probability assigned to class $k$ across all pixels predicted as class $k$. 
Higher values indicate better performance in this case.

\noindent(ii) \emph{MaNo}~\cite{xie2024mano} is adapted as it was originally designed for assessing general model performance in image classification, to evaluate class-wise performance in semantic segmentation. 
The higher the MaNo score, the better it is.

\noindent (iii) \emph{``ITI''}~\cite{hu2023mixed} is the Inter-class separability to Intra-class similarity ratio, and captures both the separability of the classes in the feature space as well as the ability to keep samples of the same class close together. For this metric, higher scores represent better results. 

\noindent These metrics are detailed in~\cref{sec:unsup-metrics}.

\noindent Overall, our entropy measure (\cf~\cref{eq:entropy}) is the most effective, although \emph{Avg. Probability} performs on par.

\begin{table}[t]
    \centering
    \small
    \resizebox{0.9\linewidth}{!}{%
    \setlength{\tabcolsep}{4pt}
    \begin{tabular}{llccc}
    \toprule
    \textbf{Fusion} & \textbf{Conflict} & \makecell{\textbf{\MASKCLIP}\\{+\method{}}} & \makecell{\textbf{\NACLIP}\\{+\method{}}} & \makecell{\textbf{\CLIPDINOISER}\\{+\method{}}} \\
    \midrule
    \multicolumn{2}{c}{Average-all} & 24.7 & 36.6 & 31.2 \\
    \midrule
    Expert & Default & 24.8 & 36.6 & 32.0 \\
    Expert & Average & 25.6 & 36.7 & 34.4 \\
    \ourrow{}Expert & Highest & \textbf{25.8} & \textbf{37.0} & \textbf{34.6} \\
    \bottomrule
    \end{tabular}%
    }
    \caption{{\textbf{Strategies for the fusion of expert predictions.} Exploring various fusion strategies on \CITYSCAPES shows the benefit of \ourcolorbox{our strategy}, which outperforms all others across models.}}
    \label{tab:fusion_strategies}
\end{table}
\begin{figure}[t]
    \centering
    \begin{minipage}[c]{0.44\linewidth}\vspace{-1em}
        \setlength{\tabcolsep}{8pt}
        \resizebox{1.0\linewidth}{!}{%
            \begin{tabular}{cc}
                \toprule
                \textbf{Metric}         & \textbf{mIoU} \\
                \midrule
                Avg. Probability        & 34.4          \\
                MaNo~\cite{xie2024mano} & 29.9          \\
                ITI                     & 30.0          \\
                \ourrow{}Entropy        & \textbf{34.6} \\
                \bottomrule
            \end{tabular}%
        }
        \subcaption{Expert metrics}\label{tab:model_comparison}
    \end{minipage}\hfill%
    \begin{minipage}[c]{0.53\linewidth}
        \includegraphics[width=1.0\linewidth]{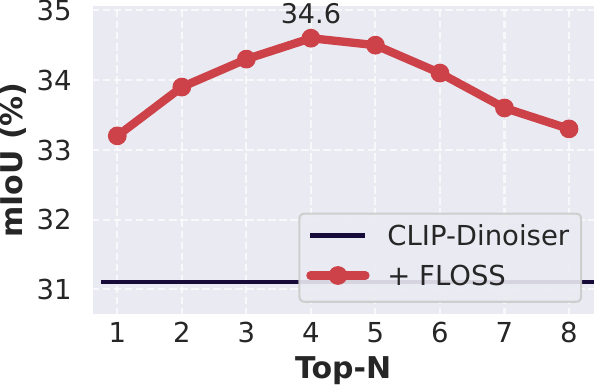}
        \subcaption{Effect of Top-N}
        \label{tab:topn_ablation}
    \end{minipage}

    \caption{\textbf{Ablation of experts.} We ablate our choices of metrics to identify experts as well as the number of experts per class. \subref{tab:model_comparison}~shows unsupervised metrics to identify the experts, showing that \ourcolorbox{our entropy} yields the best performance, although `Avg. Probability' performs on par. \subref{tab:topn_ablation}~reports performance of \method{} when varying the number of $N$ experts. Results are reported on the \CITYSCAPES dataset using the \CLIPDINOISER model.
    }
\end{figure}
\setlength{\tabcolsep}{2pt}
\begin{figure}[t]
    \centering
    \setlength{\tabcolsep}{6pt}

    \resizebox{0.56\linewidth}{!}{%
        \includegraphics[width=1.0\linewidth]{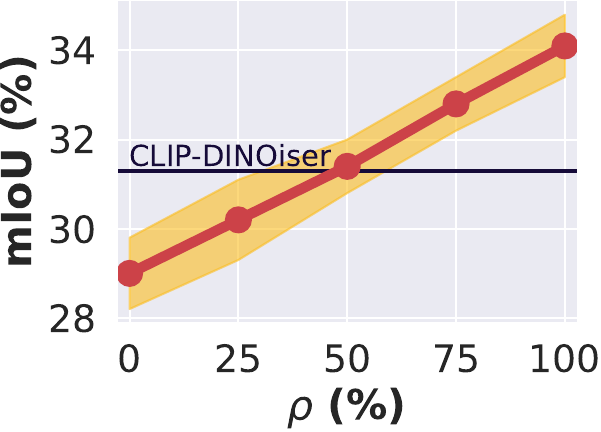}
    }

    \caption{\textbf{Effect of class-expert quality.} We report the performance of \method in the oracle-based setting, where each class-wise set of 4 experts contains a ratio ($\rho$) of true experts. The experiment is conducted on \CITYSCAPES using the \CLIPDINOISER model.}
    \label{fig:expertquality}
\end{figure}
\smallskip\noindent\textbf{Impact of class-expert identification quality.}
At the core of our method is the identification of class-experts. While we demonstrated there are templates which are class-expert, for each class there exist also under-performing templates (\cf~\cref{fig:individualtemplates}) which, if selected, could harm the performance.

To measure the impact of experts quality on \method{}, we devise \textit{an oracle-based} experiment using $N=4$ templates per-class , where we have access to ground truth labels to identify the true class-experts. We randomly sample a fixed ratio $\rho$ of templates among true experts (\textit{i.e.}, $\expert_k$ from~\cref{eq:experts}) and the remaining ones among non-experts (\textit{i.e.}, $\{\mathcal{T}_m\} \setminus \expert_k$). For example, a ratio of $\rho = 75\%$ indicates that 3 templates are experts and 1 is non-expert. The outcome of this experiment on \CITYSCAPESabbr using the \CLIPDINOISER model is reported in~\cref{fig:expertquality} and shows an expected performance boost when the ratio of true experts increases. It also highlights that $50\%$ of true experts is sufficient for \method{} to surpass the original \CLIPDINOISER. Interestingly, the performance of \CLIPDINOISER+ \method{} in \cref{tab:sota_table} (34.6) slightly surpasses the performance at \mbox{$\rho=100\%$} shown in \cref{fig:expertquality} (34.1). This however does not indicate that our estimated templates in \cref{tab:sota_table} are all true experts -- as reflected by~\cref{tab:accuracy} -- but rather that our correctly estimated experts are on average better than random true experts.

{To further measure the upper bound of \CLIPDINOISER+ \method{}, we implement an oracle version using only the \textit{best true class-expert} \mbox{(i.e., $\{\mathcal{T}_{\tilde{m}} \mid \tilde{m} \in \texttt{Top-N}\big(\Omega_k(\mW(\mathcal{T}_m))\big) \}$)}. On \CITYSCAPESabbr, the oracle with $N \in \{1, 2, 3, 4\}$ achieves respectively 37.1/37.0/37.3/37.5\%, showing the potential for improvement assuming the identification of better experts.}

\smallskip\noindent\textbf{FLOSS using the validation set.} First, we investigate the effectiveness of FLOSS in a transductive setting. Using the validation set of CS for expert selection, CLIP-DINOiser + FLOSS achieves 34.4\% mIoU on the same set, compared to 31.3\% for CLIP-DINOiser. Second, using a portion of the validation set to select experts, \textit{i.e.}, 5/25/50/75 val images, we achieve 32.8/33.0/33.0/33.3\% on the rest of the validation images, compared to 30.9\% for CLIP-DINOiser alone.

\smallskip\noindent\textbf{Impact of different backbones.}
\begin{table}[t]
    \centering
    \small
    \setlength{\tabcolsep}{4pt}
    \begin{tabular}{l|cc|c|cc|c}
    \hline
    \multirow{2}{*}{\textbf{Method}} & \multicolumn{3}{c|}{\textbf{ViT-B/16}} & \multicolumn{3}{c}{\textbf{ViT-L/14}} \\
    & \CITYSCAPESabbr & \PASCALCOFIVENINEabbr & Avg. & \CITYSCAPESabbr & \PASCALCOFIVENINEabbr & Avg. \\
    \toprule
    SCLIP~\cite{wang2024sclip}             & 32.2 & 34.2 & 33.2 & 21.3 & 25.2 & 23.3 \\
    GEM~\cite{bousselham2024grounding}               & 32.6 & 35.6 & 34.1 & 27.1 & 28.1 & 27.6 \\
    \midrule
    \NACLIP~\cite{hajimiri2025pay}          & 35.5 & 35.2 & 35.4 & 31.4 & 32.1 & 31.8 \\
    \ourrow + \method & \textbf{37.0} & \textbf{35.9} & \textbf{36.5} & \textbf{32.4} & \textbf{32.4} & \textbf{32.4} \\
    \bottomrule
    \end{tabular}
    \caption{\textbf{Effect of backbone.} Results (mIoU) show that \mbox{\NACLIP + \method} outperforms \NACLIP across ViT-B/16 and ViT-L/14 CLIP backbones.}
    \label{tab:tab_ablation_backbone}
\end{table}
~\cref{tab:tab_ablation_backbone} demonstrates the effectiveness of our proposed \method when integrated with \NACLIP using the ViT-L/14 backbone. Our approach achieves consistent improvements, reaching 32.4\% mIoU on both \CITYSCAPES and PC59 datasets, leading to a 0.6 gain in average performance over \NACLIP. These results validate the effectiveness of our method when using large-scale vision transformers.

\section{Perspectives and Future Work}
\label{sec:perspective}

Our experiments show that entropy is a strong estimator of class-experts. 
However, future research could develop even better proxies for prediction accuracy. Template-based methods appear especially promising, with significant potential for improving OVSS performance. 
As illustrated in~\cref{fig:best_perf_vs_standars}, there is still a large gap to close---for example, selecting the \textbf{best expert} for the ``sky'' class could improve IoU by over $30$ percentage points.
These ``upper-bounds'' could be further increased by using more or augmented templates beyond the $80$ currently used.

\begin{figure}[t]
    \includegraphics[width=\linewidth]     
    {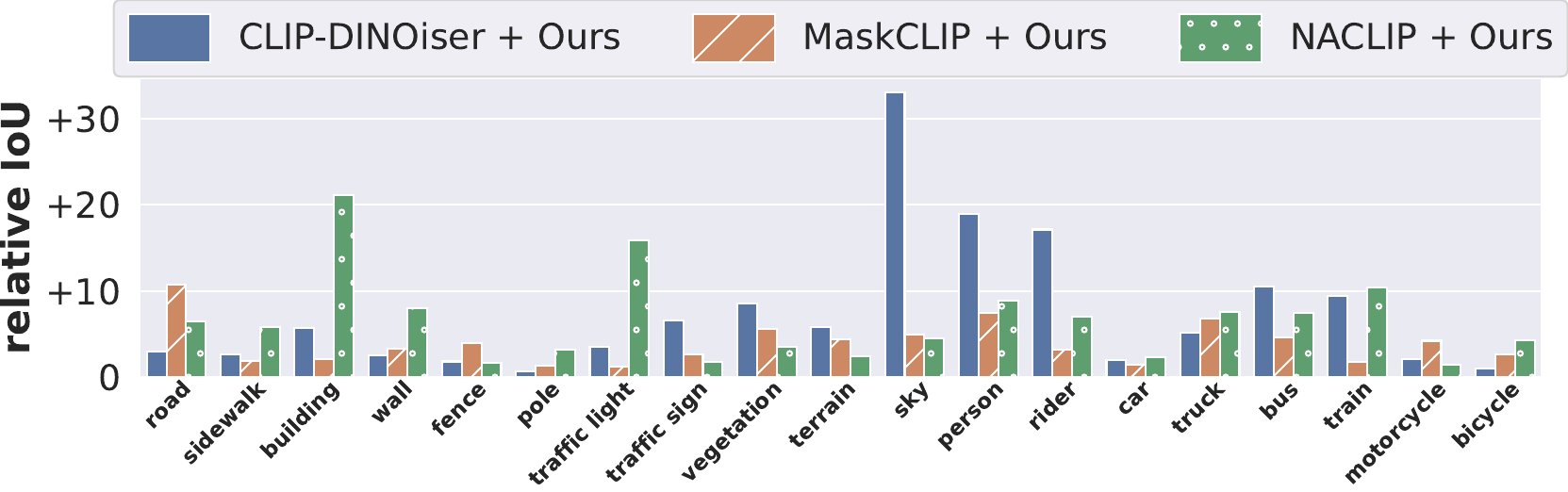}
    \caption{\textbf{Performance of best-class experts.}
    Performance gains (IoU) when using best class-experts versus standard template-averaging. Results are shown for CLIP-DINOiser, MaskCLIP and NACLIP. The bars represent the IoU difference between using best class-experts (selected with ground-truth labels) and the default template-averaging approach.}
    \label{fig:best_perf_vs_standars}
\end{figure}
\section{Conclusion}
\label{sec:conclusion}
We propose a new task for improving OVSS models without having access to labels and without training, from a prompt template perspective. Our work is motivated by an intriguing observation: for each class, some templates excel in segmenting it. We refer to these templates as class-experts. Given a set of unlabeled images, our \method uses the entropy of predictions as an unsupervised metric to identify these class-experts, thus not requiring any training or labels. Once selected, we propose a simple scheme for fusing the predictions of the class-experts, resulting in an improved overall OVSS performance in the inductive setting, even in the presence of distribution shifts. Additionally, we corroborate the effectiveness of \method, which benefits from being plug-and-play, and further show that only few unlabeled images can be sufficient for expert selection. Furthermore, the oracle performance shows the potential of the class-experts, paving the way for future research.

{\small
\smallskip\noindent\textbf{Acknowledgment.} This research was partially funded by the French Agence Nationale de la Recherche (ANR) with the project SIGHT (ANR-20-CE23-0016) and supported by ELSA - European Lighthouse on Secure and Safe AI funded by the European Union under grant agreement No. 101070617. We sincerely thank Telecom Paris for providing the resources necessary to run our experiments and Nacereddine Laddaoui for his help with infrastructure. We are also grateful to Ivan Lopes for proofreading.
\def\maketitlesupplementary
   {
   \newpage
       \twocolumn[
        \centering
        
         \Large\textbf{Supplementary Material}
         \vspace{2.0em}
        
       ]
   }
   
\maketitlesupplementary

\appendix
\renewcommand{\appendixautorefname}{Section}

\section{Additional experiments}
In this section, we provide additional experimental results and analyses that complement our main findings. First, we showcase the consistent improvements brought by our method across different models and datasets (Section~\ref{sec:improvements}). We also report the computational overhead analysis including inference time and GPU memory requirements (Section~\ref{sec:inf_gpu}). Then we extend our empirical analysis of class-expert templates to additional datasets and models (Section~\ref{sec:empirical-obs}). We follow with a detailed assessment of our expert identification method's accuracy (Section~\ref{sec:supp-expert-accuracy}) and an in-depth explanation of the different unsupervised metrics we investigated (Section~\ref{sec:unsup-metrics}). We present a comprehensive study of the relationship between entropy and IoU across different settings (Section~\ref{sec:entropy-iou}). For reproducibility, we provide the complete list of ImageNet templates used across all experiments (Section~\ref{sec:imagenet-templates}). Finally, we provide qualitative visual results that demonstrate the effectiveness of our class-expert fusion approach (Section~\ref{sec:qualitative-results}).

\subsection{Performance improvements across models}
\label{sec:improvements}
\cref{fig:example_figure} provides a graphical view across datasets of the improvement in mIoU when plugging \method{} on existing OVSS models. This demonstrates the general applicability of our approach, which can be seamlessly integrated with existing models without requiring any additional training.

\subsection{Inference time and GPU memory}
\label{sec:inf_gpu}
We provide the results of the inference time and GPU memory of \method{} when it is plugged to existing open-vocabulary models across different datasets. As shown in~\cref{tab:tab_inf_gpu}, the computational overhead varies significantly based on the number of classes in each dataset. For datasets with a smaller number of classes such as \CITYSCAPESabbr~\cite{cordts2016cityscapes}~(19 classes) and \PASCALVOCTWENTYabbr~\cite{pascal-voc-2012}~(20 classes), \method{} incurs minimal computational overhead, with modest increases in both inference time and GPU memory usage. However, for datasets with a larger number of classes like \PASCALCOFIVENINEabbr~\cite{mottaghi2014role}~(59 classes) and \ADETWENTYKabbr~\cite{zhou2019semantic}~(150 classes), the overhead becomes more substantial, with notable increases in both inference time and peak GPU memory consumption, reflecting the scaling nature of cosine similarity computations across class-experts. It is worth noting that \NACLIP~\cite{hajimiri2025pay} exhibits particularly high inference times on \CITYSCAPESabbr~for both the baseline and \method{}, which stems from the preprocessing pipeline in the original \NACLIP~implementation that resizes \CITYSCAPESabbr~images to 1120$\times$560 resolution and employs sliding window inference with 224$\times$224 crops and a stride of 112, requiring 36 overlapping patch inferences to process each complete image.
\begin{table}[h]
    \centering
    \setlength{\tabcolsep}{4pt}
    \resizebox{\linewidth}{!}{%
        \begin{tabular}{l|cc|cc|cc|cc}
            \toprule
            \multirow{2}{*}{\textbf{Method}} & \multicolumn{2}{c}{\textbf{\CITYSCAPESabbr}} & \multicolumn{2}{c}{\textbf{\PASCALVOCTWENTYabbr}} & \multicolumn{2}{c}{\textbf{\PASCALCOFIVENINEabbr}} & \multicolumn{2}{c}{\textbf{\ADETWENTYKabbr}}                                                                                   \\
                                             & \textbf{Inf. time}                           & \textbf{GPU mem.}                                 & \textbf{Inf. time}                                 & \textbf{GPU mem.}                            & \textbf{Inf. time} & \textbf{GPU mem.} & \textbf{Inf. time} & \textbf{GPU mem.} \\
            \midrule
            \CLIPDINOISER                    & 31                                           & 1203/1274                                         & 23                                                 & 1184/1263                                    & 24                 & 1204/1397         & 23                 & 1252/1661         \\
            + FLOSS                          & 50                                           & 1204/1940                                         & 35                                                 & 1184/2034                                    & 83                 & 1211/7225         & 339                & 1295/19249        \\
            \midrule
            \MASKCLIP                        & 30                                           & 1195/1259                                         & 23                                                 & 1175/1251                                    & 23                 & 1195/1436         & 22                 & 1242/1652         \\
            + FLOSS                          & 59                                           & 1196/1899                                         & 43                                                 & 1175/1986                                    & 108                & 1202/7142         & 421                & 1286/19240        \\
            \midrule
            \NACLIP                          & 100                                          & 449/571                                           & 18                                                 & 365/408                                      & 23                 & 373/458           & 22                 & 422/739           \\
            + FLOSS                          & 249                                          & 449/571                                           & 41                                                 & 366/1232                                     & 91                 & 377/1092          & 274                & 444/4854          \\
            \bottomrule
        \end{tabular}}
    \caption{\textbf{Inference time and GPU memory analysis.} We report inference time (ms) and GPU memory usage (current/peak in MB) for different OVSS models with and without \method{} across four datasets. }
    \label{tab:tab_inf_gpu}
\end{table}

\subsection{Empirical observations on other settings}
\label{sec:empirical-obs}
The foundation of our work is the observation that there exist class-wise expert templates. We extend here our initial analysis to 4 datasets, for both \CLIPDINOISER~\cite{wysoczanska2024clip} in \cref{fig:individualtemplates_clipdinoiser} and \NACLIP in  \cref{fig:individualtemplates_naclip}, showing that our initial observations hold in these settings.
Interestingly, we note that for \PASCALCOFIVENINE, ``template-averaging'' performs already very well and most individual template underperform it. Arguably, this might be related to the proximity of \PASCALVOCTWENTY with ImageNet for which the templates of CLIP were engineered~\cite{radford2021learning}.

\subsection{Quality of experts}
\label{sec:supp-expert-accuracy}
We report the quality of our estimated \mbox{top-4} experts using entropy as unsupervised metric for both \CLIPDINOISER in \cref{fig:topn_acc_clipdinoiser} and \NACLIP in \cref{fig:topn_acc_naclip}.
In details, for each class $k \in [1, K]$, the quality is computed as the normalized intersection between the set of estimated experts $\estexpert{}_k$ and the true set of experts $\expert_k$, as detailed in~\cref{eq:experts_accuracy}.
For each dataset, we also report the average accuracy over all classes, \ie, $\frac{1}{K} \sum_k \hat{\rho}_k$, shown as a horizontal line.

While we observe a high variability of the quality across classes, our average top-4 quality is around 50\%, meaning that half of our estimated expert templates are usually actual class-experts. We also note that the identification of experts on \PASCALVOCTWENTY is less accurate, which may stem from the rare class-experts on said dataset and the high performance of the ``template-averaging'' model.

\subsection{Unsupervised metrics for expert identification}
\label{sec:unsup-metrics}
In~\cref{sec:ablation}, we compared entropy to three other metrics for expert identification. We provide here details about these metrics.
Given a single-template classifier $\mW(\mathcal{T}_m)$ making predictions on unlabeled images, (i) \emph{Avg. Probability} computes the average probability of all pixels predicted as class $k$ by $\mW(\mathcal{T}_m)$. It writes:
\begin{equation}
    \text{Avg.Prob}_k(\mathcal{T}_m) = \frac{1}{C_{m,k}} \sum_{i=1}^{C_{m,k}}\text{softmax}(\mathbf{q}_{i})_{k}
\end{equation}

\noindent (ii) \emph{MaNo} is based on low density separation (LDS) assumption.
The LDS principle suggests a direct correlation between the magnitude of logit values and a model's generalization capabilities, so the higher the MaNo score, the better it is. For each class $k$, we compute the average $L_p$ norm of the probability vectors for pixels predicted as that class:
\begin{equation}
    \text{MaNo}_k(\mathcal{T}_m) = \left( \frac{1}{C_{m,k}K} \sum_{i=1}^{C_{m,k}}\sum_{k=1}^{K} \left| \text{softrun}(\mathbf{q}_{i})_{k} \right|^{p} \right)^{\frac{1}{p}},
\end{equation}
where $\text{softrun}$ is the normalization function used to avoid error accumulation, and we use $p=4$ in practice, as done in the original paper.

\noindent (iii) \emph{Inter-to-Intra (ITI)} is a class ratio that quantifies feature separability by examining the relationship between inter-class separation and intra-class compactness for each class $k$. The intra-class compactness measure $D_{\text{intra,k}}(\mathcal{T}_m)$ calculates the average squared Euclidean distance between the global class feature centroid $\bar{f}_k$ and the class-specific features $\bar{f}^{i}_k$ (denoting the average feature for class $k$ in image $i$) across all $N$ training images, capturing feature variability within the class.

\noindent Conversely, the inter-class separability measure $D_{\text{inter,k}}(\mathcal{T}_m)$ computes the average squared distance between the target class centroid $\bar{f}_k$ and all other class centroids $\bar{f}_i$, measuring how distinctly a class is represented in feature space. The ITI ratio $\text{ITI}_k(\mathcal{T}_m)$ then provides a normalized metric where higher values indicate well-separated and compact feature representations, with $\bar{f}_k$ representing the class $k$'s feature centroid across the entire dataset, $N$ representing the training dataset size, and $K$ signifying the total number of semantic classes.

\noindent $D_{\text{intra,k}}(\mathcal{T}_m)$, $D_{\text{inter,k}}(\mathcal{T}_m)$ and $\text{ITI}_k(\mathcal{T}_m)$ write as:
\begin{align}
    D_{\text{intra,k}}(\mathcal{T}_m) & = \frac{1}{N}\sum_{i=1}^{N} \|\bar{f}_k - \bar{f}^{i}_k\|_2^2 \notag                \\
    D_{\text{inter,k}}(\mathcal{T}_m) & = \frac{1}{N-1} \sum_{i=1}^{N-1} \|\bar{f}_k - \bar{f}_i\|_2^{2}                    \\
    \text{ITI}_k(\mathcal{T}_m)       & = \frac{D_{\text{inter,k}}(\mathcal{T}_m)}{D_{\text{intra,k}}(\mathcal{T}_m)}\notag
\end{align}

\subsection{Studying the relation of entropy and IoU}
\label{sec:entropy-iou}
We analyze the relationship between entropy and IoU across different models and datasets. \cref{fig:entropy_dinoiser_cs,fig:entropy_dinoiser_pco} show this relationship for \CLIPDINOISER, while \cref{fig:entropy_naclip_cs,fig:entropy_naclip_pco} present the same analysis for \NACLIP. For each class, we plot the entropy and IoU of individual templates (colored dots) against the performance of the original CLIP ``template-averaging'' (dotted line). The number of valid templates (those predicting at least one pixel for the class) is indicated in parentheses next to each class name.

\subsection{ImageNet templates}
\label{sec:imagenet-templates}
For reproducibility, we provide the complete list of 80 ImageNet templates used consistently across all our experiments. These templates are applied in the exact order listed below, ensuring consistency across all reported results:
\begin{enumerate}[start=0]
    \item 'a bad photo of a \{\}.'
    \item 'a photo of many \{\}.'
    \item 'a sculpture of a \{\}.'
    \item 'a photo of the hard to see \{\}.'
    \item 'a low resolution photo of the \{\}.'
    \item 'a rendering of a \{\}.'
    \item 'graffiti of a \{\}.'
    \item 'a bad photo of the \{\}.'
    \item 'a cropped photo of the \{\}.'
    \item 'a tattoo of a \{\}.'
    \item 'the embroidered \{\}.'
    \item 'a photo of a hard to see \{\}.'
    \item 'a bright photo of a \{\}.'
    \item 'a photo of a clean \{\}.'
    \item 'a photo of a dirty \{\}.'
    \item 'a dark photo of the \{\}.'
    \item 'a drawing of a \{\}.'
    \item 'a photo of my \{\}.'
    \item 'the plastic \{\}.'
    \item 'a photo of the cool \{\}.'
    \item 'a close-up photo of a \{\}.'
    \item 'a black and white photo of the \{\}.'
    \item 'a painting of the \{\}.'
    \item 'a painting of a \{\}.'
    \item 'a pixelated photo of the \{\}.'
    \item 'a sculpture of the \{\}.'
    \item 'a bright photo of the \{\}.'
    \item 'a cropped photo of a \{\}.'
    \item 'a plastic \{\}.'
    \item 'a photo of the dirty \{\}.'
    \item 'a jpeg corrupted photo of a \{\}.'
    \item 'a blurry photo of the \{\}.'
    \item 'a photo of the \{\}.'
    \item 'a good photo of the \{\}.'
    \item 'a rendering of the \{\}.'
    \item 'a \{\} in a video game.'
    \item 'a photo of one \{\}.'
    \item 'a doodle of a \{\}.'
    \item 'a close-up photo of the \{\}.'
    \item 'a photo of a \{\}.'
    \item 'the origami \{\}.'
    \item 'the \{\} in a video game.'
    \item 'a sketch of a \{\}.'
    \item 'a doodle of the \{\}.'
    \item 'a origami \{\}.'
    \item 'a low resolution photo of a \{\}.'
    \item 'the toy \{\}.'
    \item 'a rendition of the \{\}.'
    \item 'a photo of the clean \{\}.'
    \item 'a photo of a large \{\}.'
    \item 'a rendition of a \{\}.'
    \item 'a photo of a nice \{\}.'
    \item 'a photo of a weird \{\}.'
    \item 'a blurry photo of a \{\}.'
    \item 'a cartoon \{\}.'
    \item 'art of a \{\}.'
    \item 'a sketch of the \{\}.'
    \item 'a embroidered \{\}.'
    \item 'a pixelated photo of a \{\}.'
    \item 'itap of the \{\}.'
    \item 'a jpeg corrupted photo of the \{\}.'
    \item 'a good photo of a \{\}.'
    \item 'a plushie \{\}.'
    \item 'a photo of the nice \{\}.'
    \item 'a photo of the small \{\}.'
    \item 'a photo of the weird \{\}.'
    \item 'the cartoon \{\}.'
    \item 'art of the \{\}.'
    \item 'a drawing of the \{\}.'
    \item 'a photo of the large \{\}.'
    \item 'a black and white photo of a \{\}.'
    \item 'the plushie \{\}.'
    \item 'a dark photo of a \{\}.'
    \item 'itap of a \{\}.'
    \item 'graffiti of the \{\}.'
    \item 'a toy \{\}.'
    \item 'itap of my \{\}.'
    \item 'a photo of a cool \{\}.'
    \item 'a photo of a small \{\}.'
    \item 'a tattoo of the \{\}.'
\end{enumerate}

\subsection{Qualitative results}
\label{sec:qualitative-results}
In this section, we provide qualitative visual results to complement our quantitative analysis. \cref{fig:qualitative_results} presents example segmentation results demonstrating the effectiveness of our class-expert fusion approach when applied to \CLIPDINOISER on \CITYSCAPES images. The visualizations clearly illustrate how \method{} effectively combines predictions from different class-experts, each contributing their specialized knowledge for their respective classes of expertise. Notably, we observe the significant impact of class-specific experts, such as the sky-expert, on the final fused prediction, showcasing substantial improvements in segmentation quality compared to the baseline \CLIPDINOISER predictions.

\begin{figure}[h!]
    \centering
    \includegraphics[width=\linewidth]{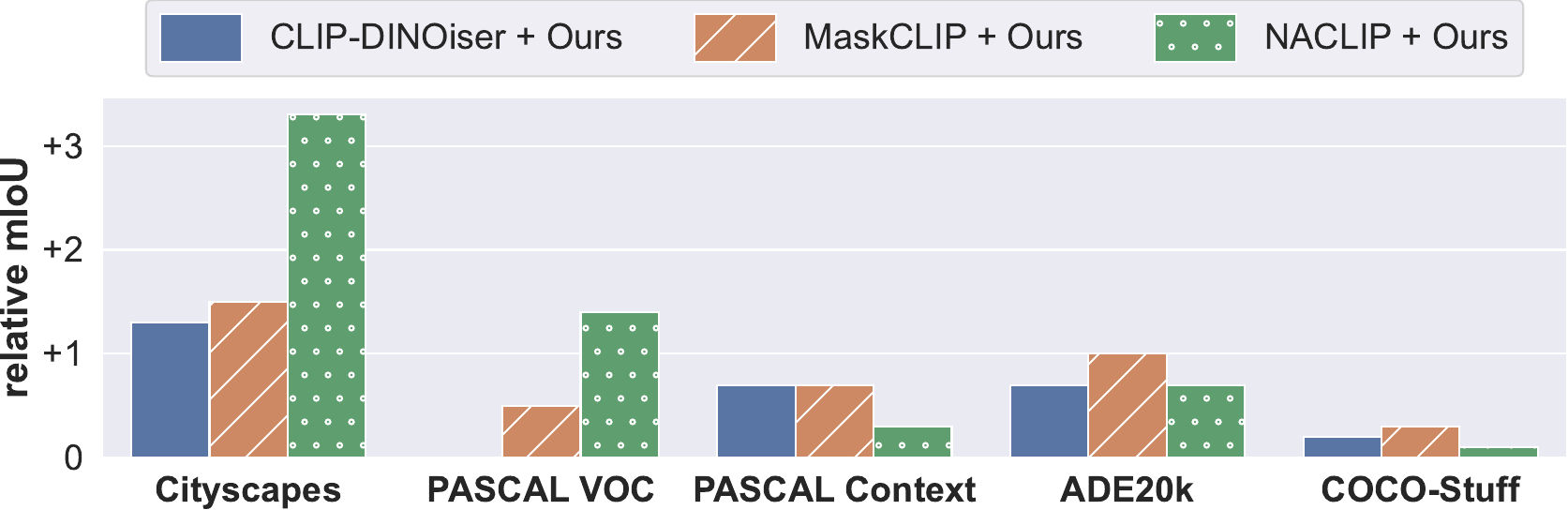}
    \caption{\textbf{Performance boost with \method{}}. We report the mIoU difference when using \method{} on top of an existing OVSS model. Our training-free method consistently improves all OVSS models across different datasets through class-expert template identification.}
    \label{fig:example_figure}
\end{figure}

\begin{figure*}
    \centering
    \begin{subfigure}[m]{1.0\linewidth}\centering
        \includegraphics[height=2cm]{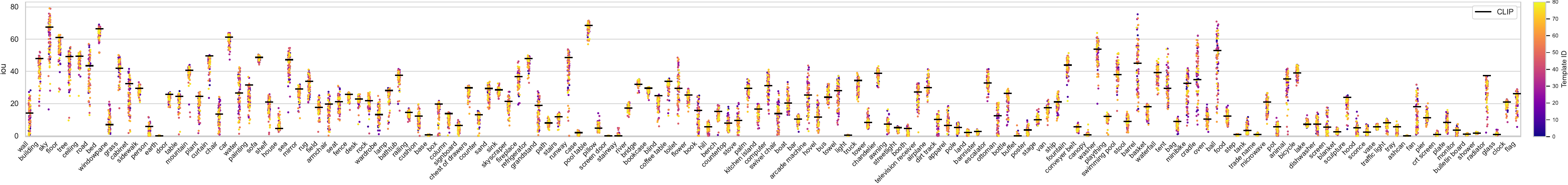}
        \caption{\ADETWENTYK}
    \end{subfigure}
    \begin{subfigure}[m]{0.25\linewidth}\centering
        \includegraphics[height=2cm]{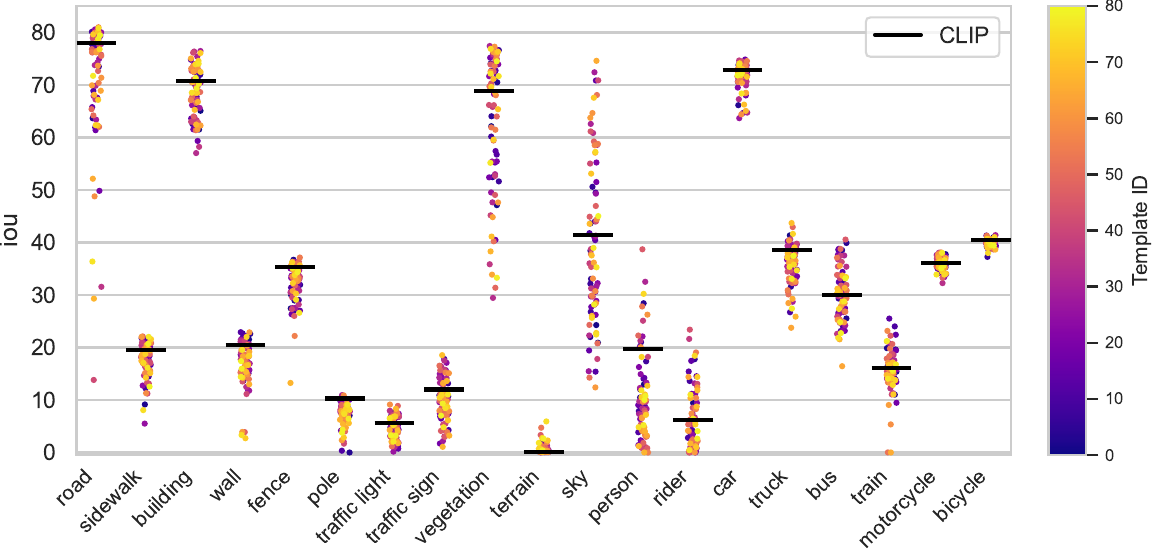}
        \caption{\CITYSCAPES}
    \end{subfigure}\hfill%
    \begin{subfigure}[m]{0.47\linewidth}\centering
        \includegraphics[height=2cm]{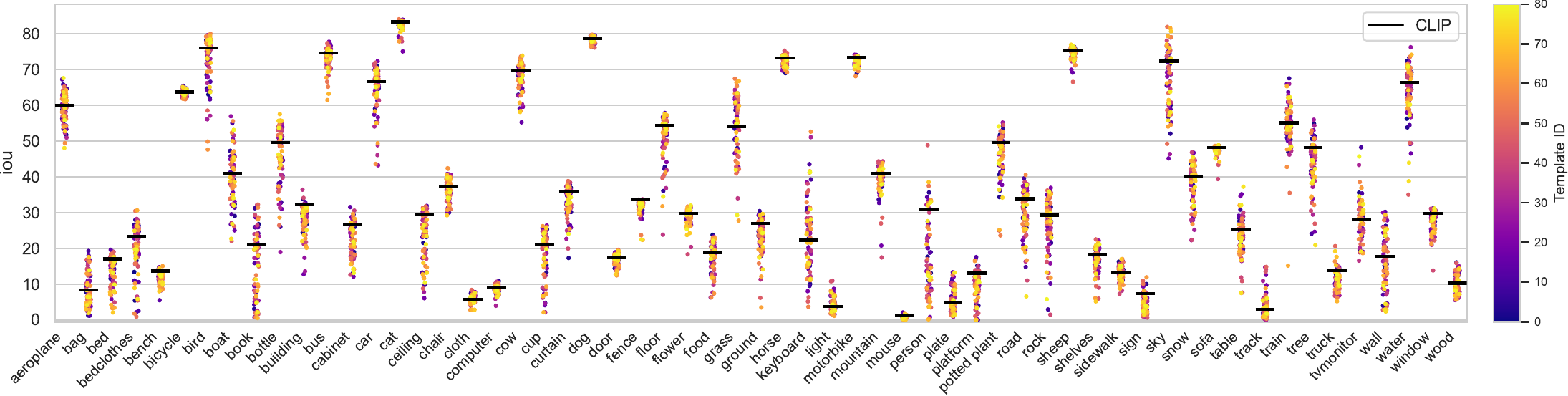}
        \caption{\PASCALCOFIVENINE}
    \end{subfigure}\hfill%
    \begin{subfigure}[m]{0.27\linewidth}\centering
        \includegraphics[height=2cm]{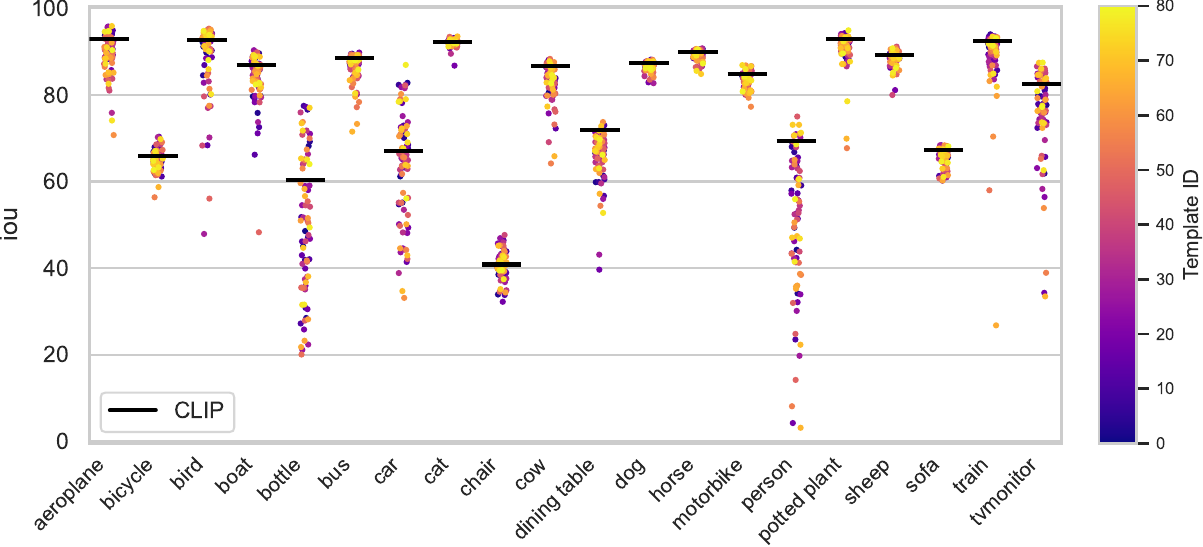}
        \caption{\PASCALVOCTWENTY}
    \end{subfigure}
    \caption{\textbf{Individual template vs average templates (\CLIPDINOISER).} }
    \label{fig:individualtemplates_clipdinoiser}
\end{figure*}
\begin{figure*}
    \centering
    \begin{subfigure}[m]{1.0\linewidth}\centering
        \includegraphics[height=2cm]{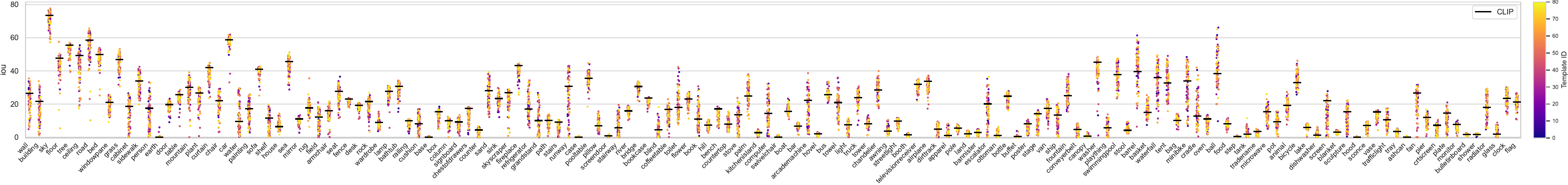}
        \caption{\ADETWENTYK}
    \end{subfigure}
    \begin{subfigure}[m]{0.25\linewidth}\centering
        \includegraphics[height=2cm]{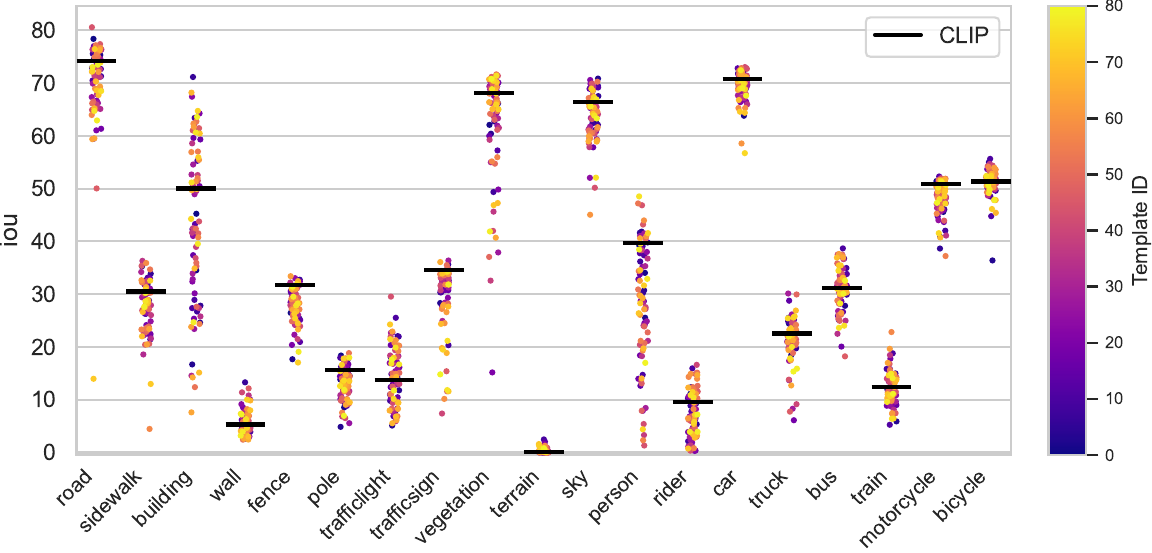}
        \caption{\CITYSCAPES}
    \end{subfigure}\hfill%
    \begin{subfigure}[m]{0.47\linewidth}\centering
        \includegraphics[height=2cm]{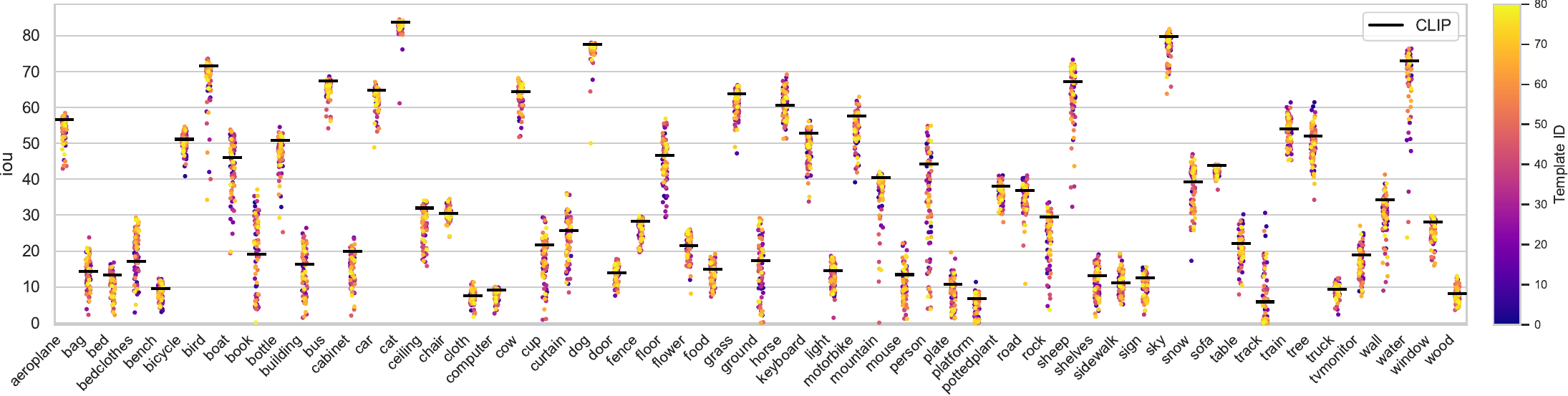}
        \caption{\PASCALCOFIVENINE}
    \end{subfigure}\hfill%
    \begin{subfigure}[m]{0.27\linewidth}\centering
        \includegraphics[height=2cm]{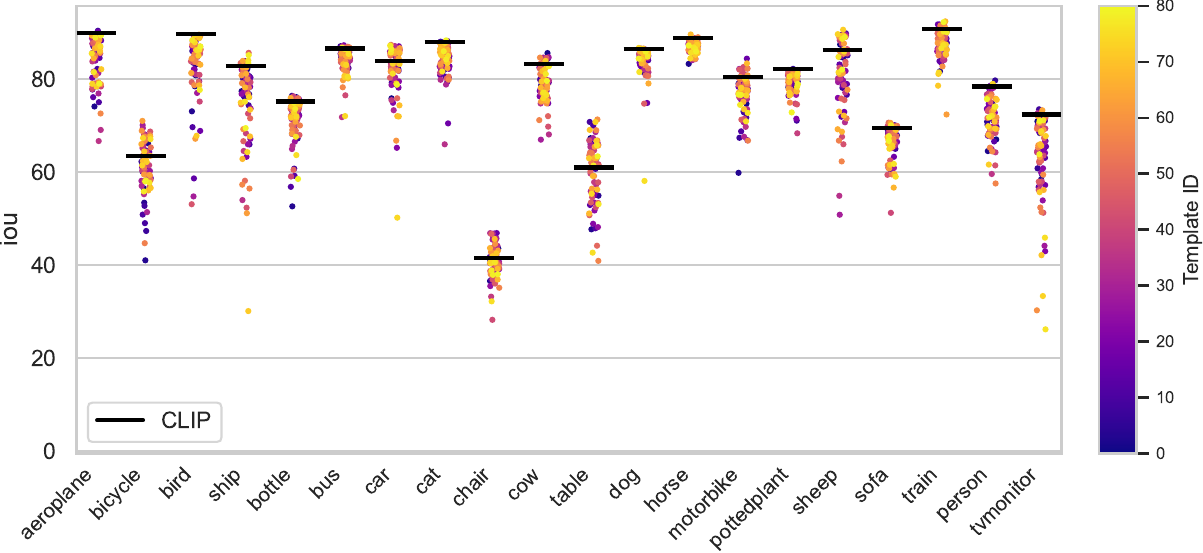}
        \caption{\PASCALVOCTWENTY}
    \end{subfigure}
    \caption{\textbf{Individual template vs average templates (\NACLIP).} }
    \label{fig:individualtemplates_naclip}
\end{figure*}

\begin{figure*}
    \centering
    \begin{subfigure}[m]{1.0\linewidth}\centering
        \includegraphics[height=2cm]{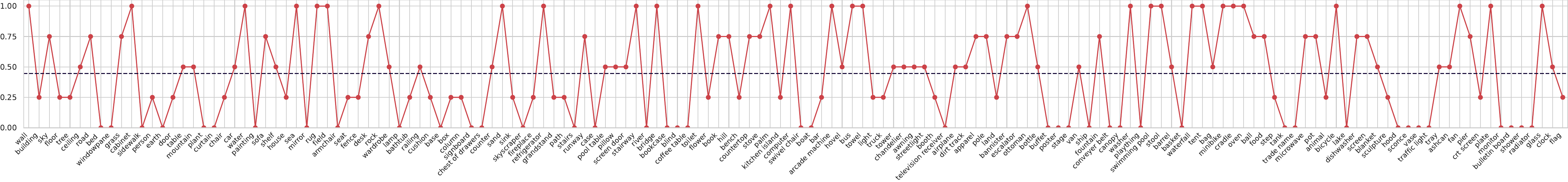}
        \caption{\ADETWENTYK}
    \end{subfigure}
    \begin{subfigure}[m]{0.25\linewidth}\centering
        \includegraphics[height=2cm]{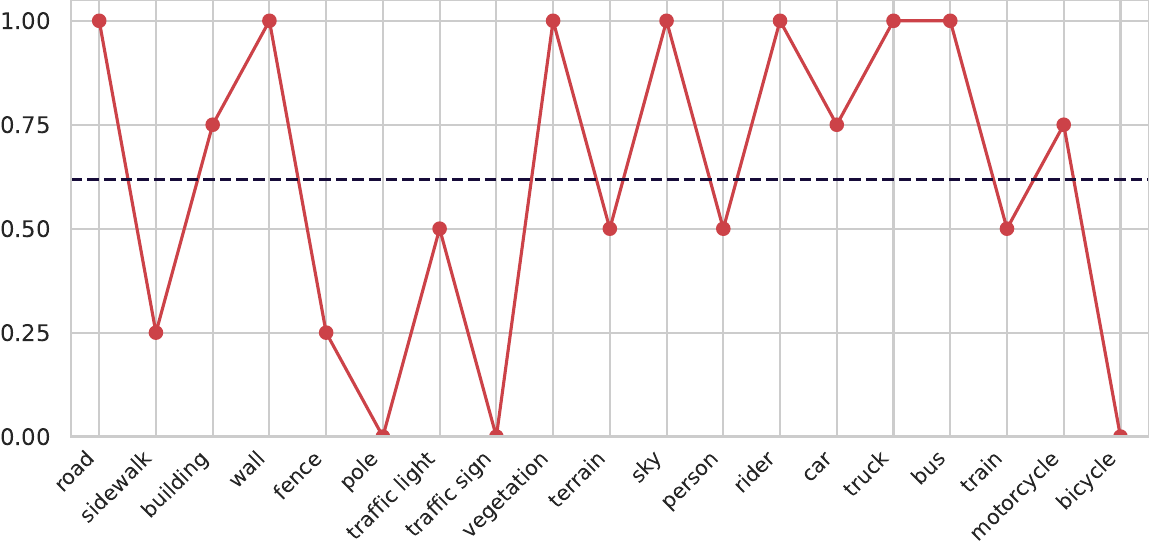}
        \caption{\CITYSCAPES}
    \end{subfigure}\hfill%
    \begin{subfigure}[m]{0.47\linewidth}\centering
        \includegraphics[height=2cm]{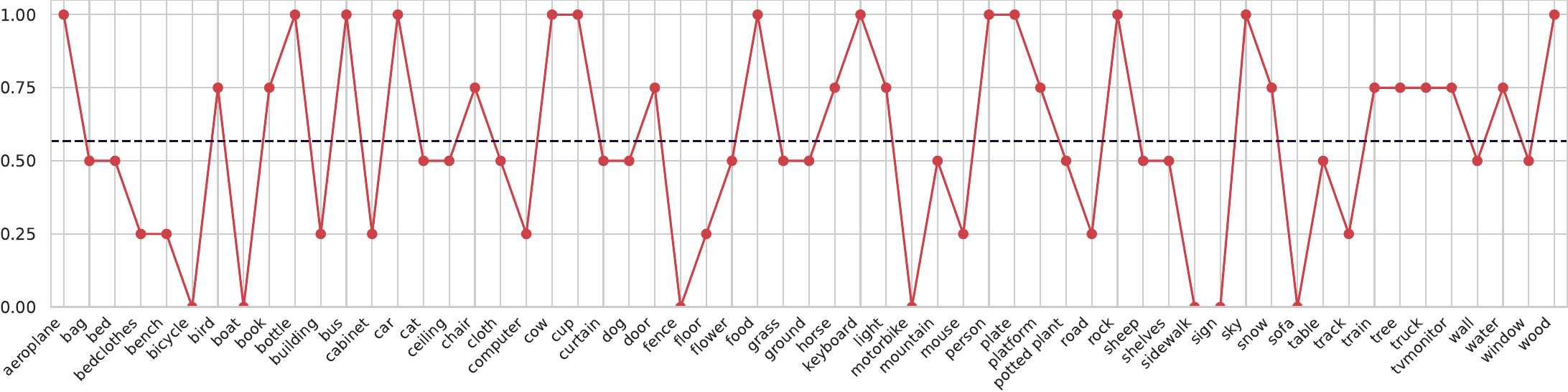}
        \caption{\PASCALCOFIVENINE}
    \end{subfigure}\hfill%
    \begin{subfigure}[m]{0.27\linewidth}\centering
        \includegraphics[height=2cm]{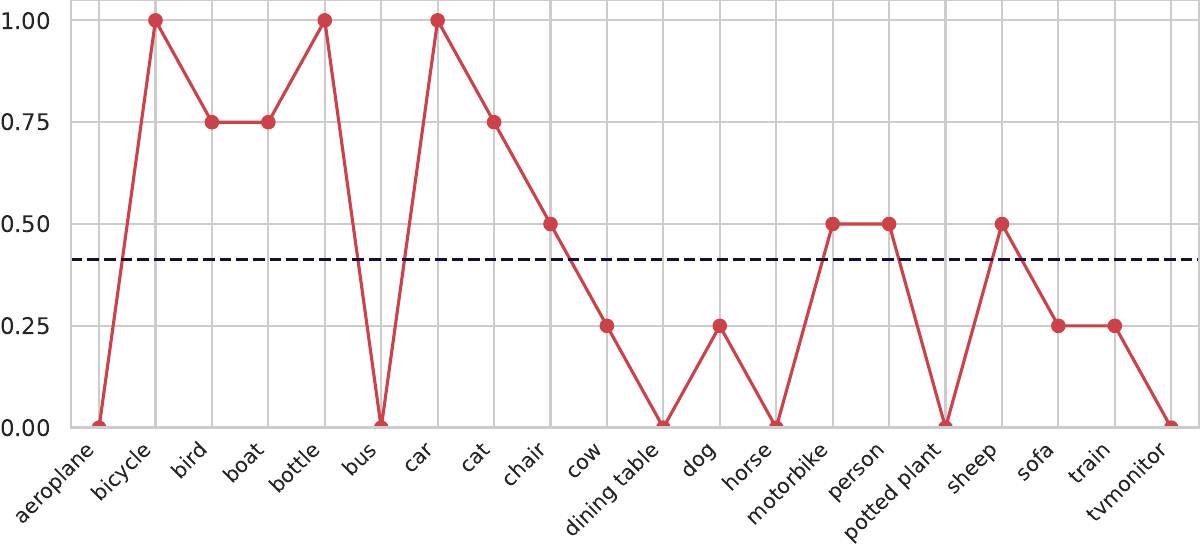}
        \caption{\PASCALVOCTWENTY}
    \end{subfigure}
    \caption{\textbf{Quality of the estimated templates (\CLIPDINOISER).} For each class $k$, we report the accuracy of the estimated top-4 experts $\hat{\expert}_k$ as the normalized intersection with the set of true experts from that class, i.e., ${\expert}_k$. The dash line indicates the average across classes.}
    \label{fig:topn_acc_clipdinoiser}
\end{figure*}
\begin{figure*}
    \centering
    \begin{subfigure}[m]{1.0\linewidth}\centering
        \includegraphics[height=2cm]{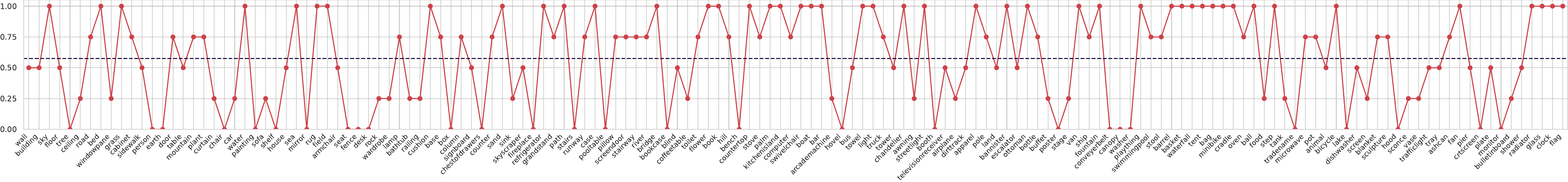}
        \caption{\ADETWENTYK}
    \end{subfigure}
    \begin{subfigure}[m]{0.25\linewidth}\centering
        \includegraphics[height=2cm]{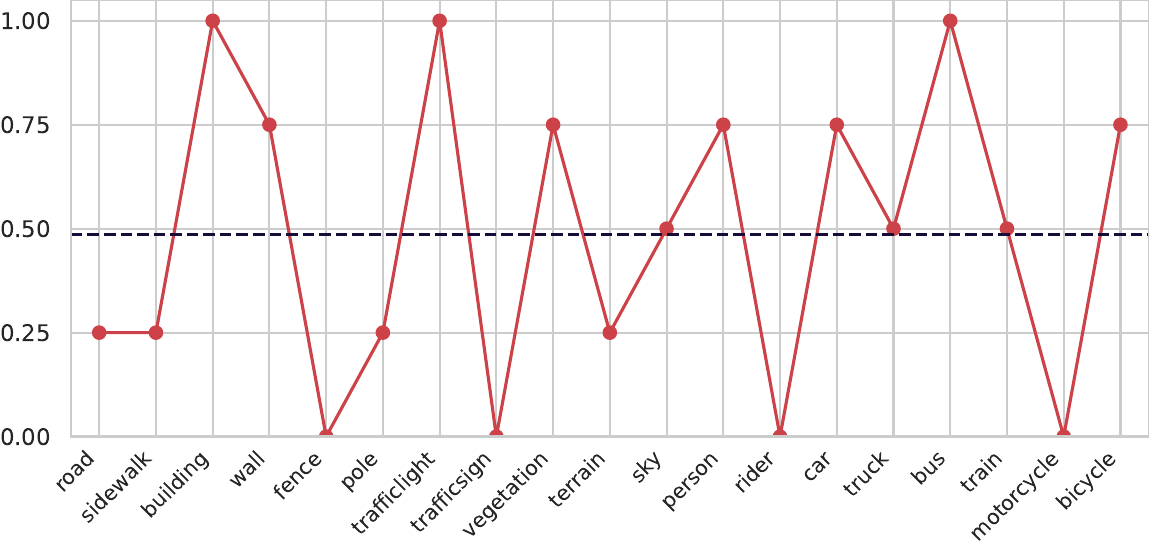}
        \caption{\CITYSCAPES}
    \end{subfigure}\hfill%
    \begin{subfigure}[m]{0.47\linewidth}\centering
        \includegraphics[height=2cm]{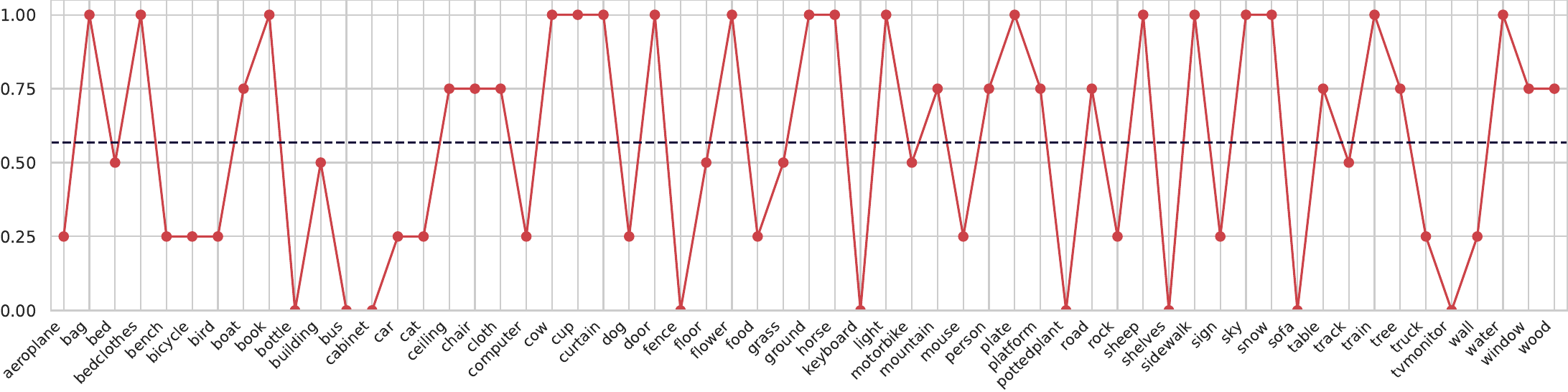}
        \caption{\PASCALCOFIVENINE}
    \end{subfigure}\hfill%
    \begin{subfigure}[m]{0.27\linewidth}\centering
        \includegraphics[height=2cm]{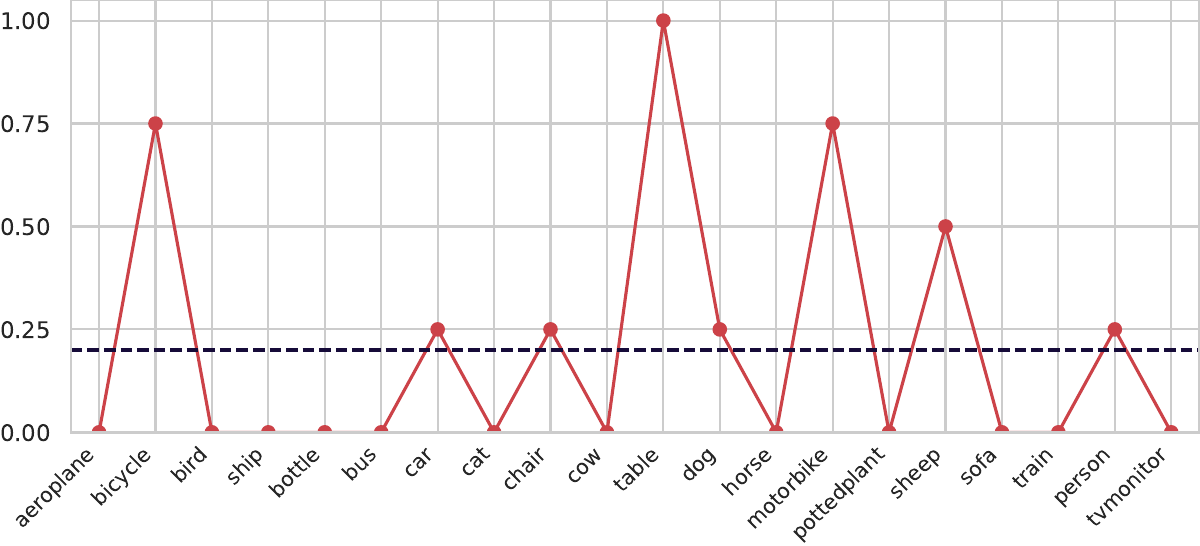}
        \caption{\PASCALVOCTWENTY}
    \end{subfigure}
    \caption{\textbf{Quality of the estimated templates (\NACLIP).} For each class $k$, we report the accuracy of the estimated top-4 experts $\hat{\expert}_k$ as the normalized intersection with the set of true experts from that class, i.e., ${\expert}_k$. The dash line indicates the average across classes.}
    \label{fig:topn_acc_naclip}
\end{figure*}

\makeatletter
\newcommand{\plotentropy}[4]{%
    \begin{figure*}%
        \includegraphics[width=\linewidth]{figures/analyses/#4}%
        \caption{\textbf{Entropy and IoU per template (#1, #2).} Each plot reports, for a given class, the entropy and IoU of all individual template (colored dots) as well as the original CLIP averaged-templates performance (dotted line).
            Note that we consider templates valid only if they predict more than 0 pixels for that class, and therefore report the number of valid templates in parenthesis next to the class name.}%
        \label{fig:#3}%
    \end{figure*}%
}
\makeatother

\plotentropy{\CLIPDINOISER}{\CITYSCAPES}{entropy_dinoiser_cs}{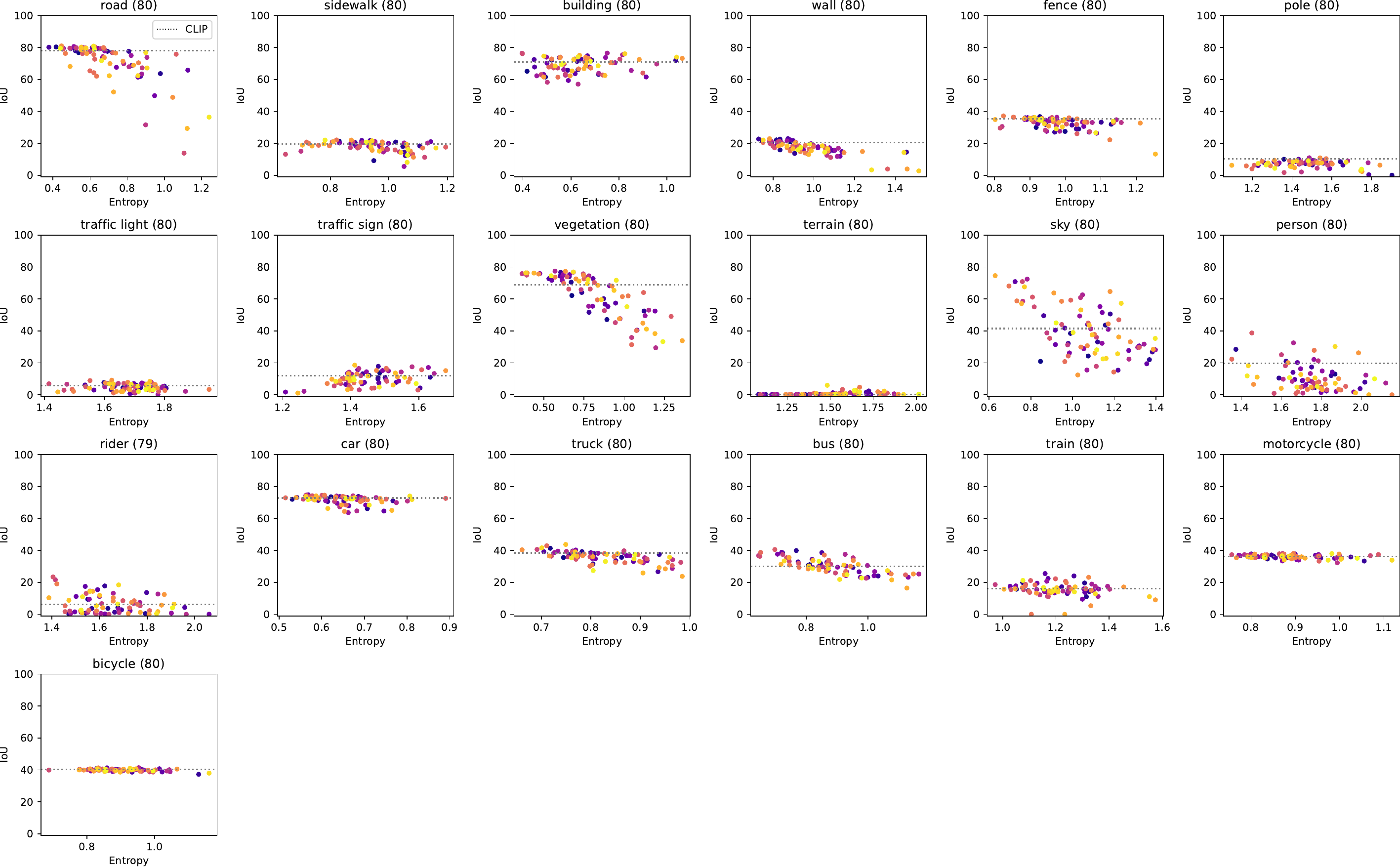}
\plotentropy{\CLIPDINOISER}{\PASCALCOFIVENINE}{entropy_dinoiser_pco}{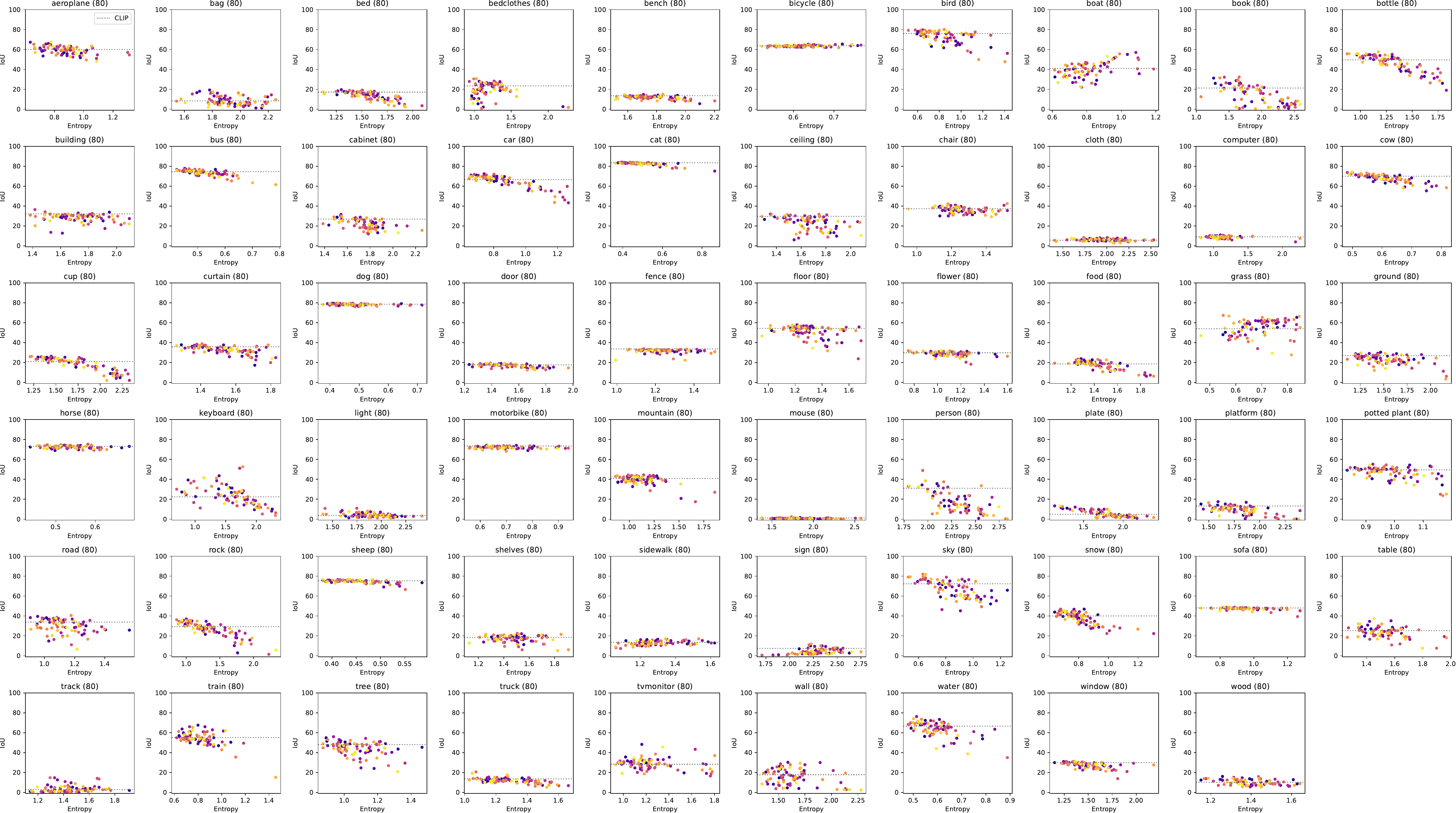}

\plotentropy{\NACLIP}{\CITYSCAPES}{entropy_naclip_cs}{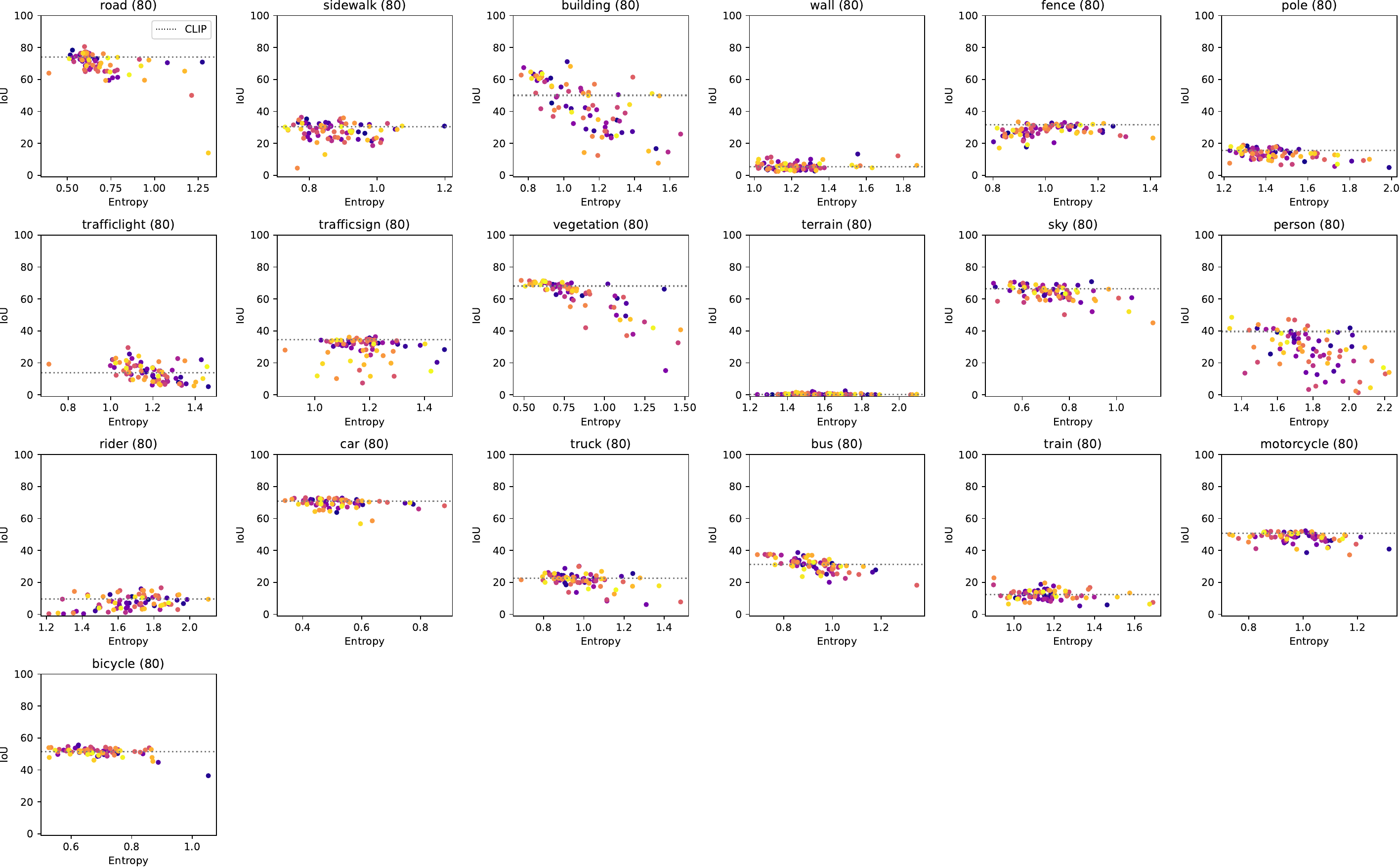}
\plotentropy{\NACLIP}{\PASCALCOFIVENINE}{entropy_naclip_pco}{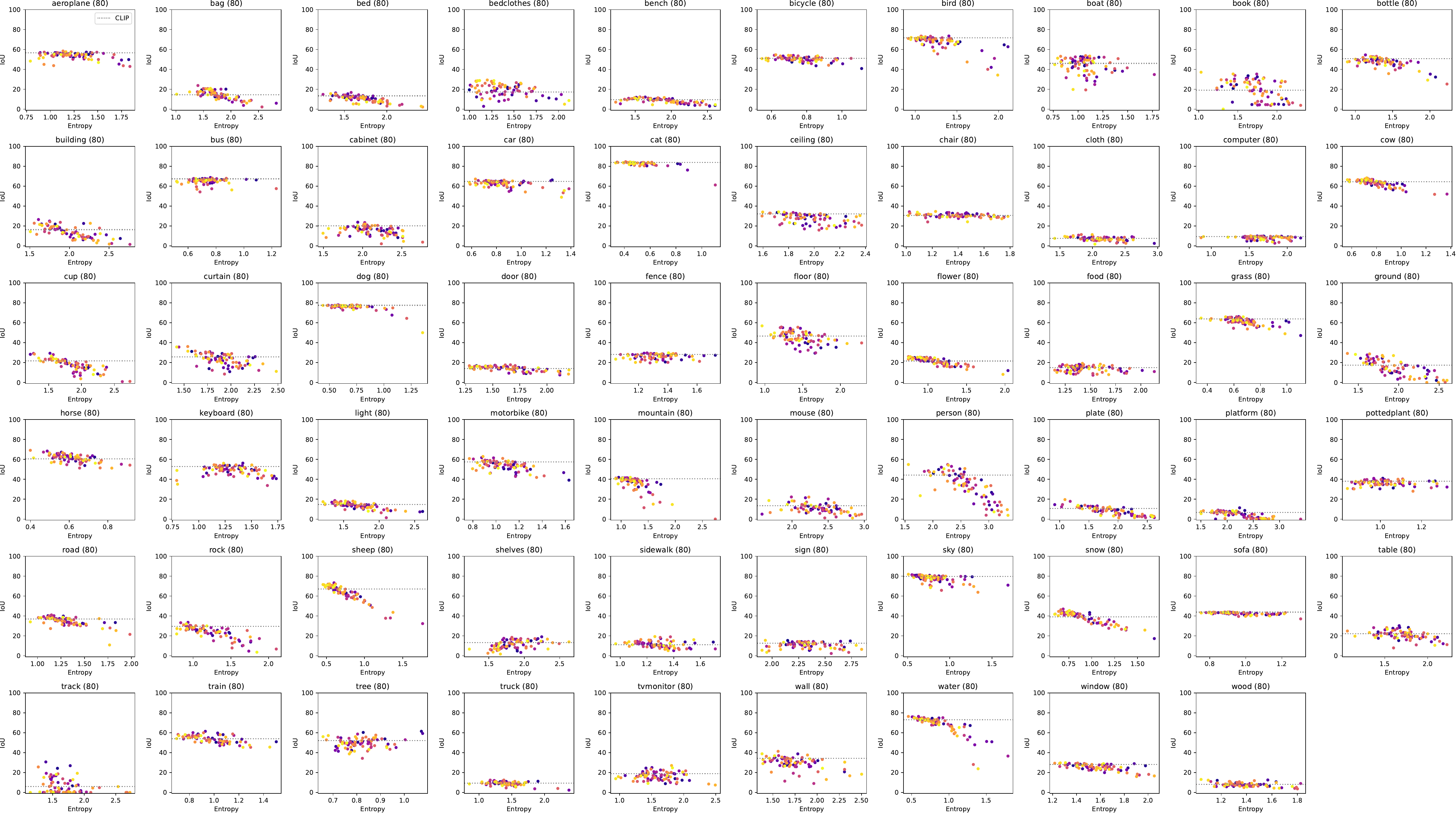}

\begin{figure*}[!t]
    \centering
    \includegraphics[width=\linewidth]{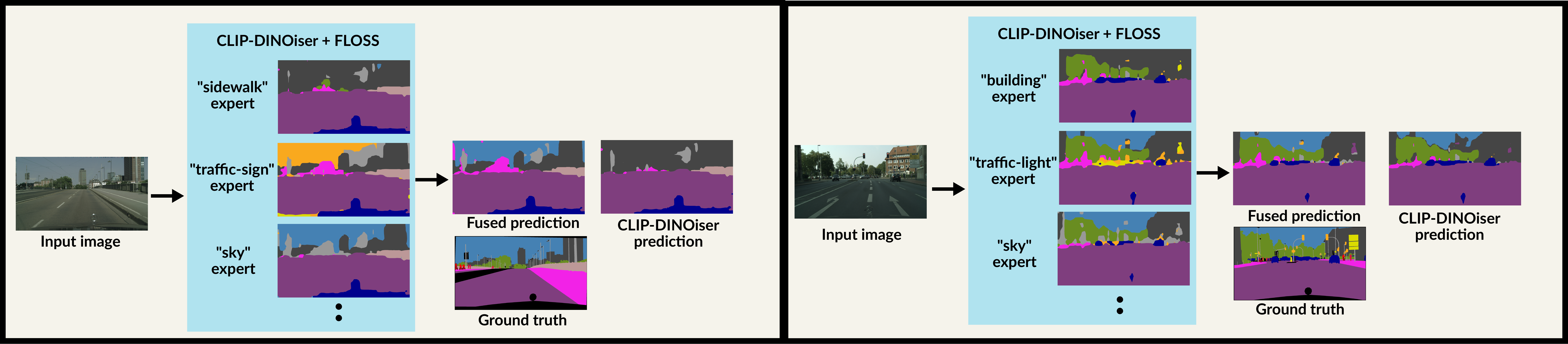}
    \caption{\textbf{Qualitative results on \CITYSCAPES.} The figure shows qualitative results displaying predictions from individual class-experts when integrating \method{} with \CLIPDINOISER on \CITYSCAPES images. The visualization demonstrates how the fused prediction effectively combines the predictions made by class-experts within their respective domains of expertise, outperforming the baseline \CLIPDINOISER prediction.}
    \label{fig:qualitative_results}
\end{figure*}
\clearpage
{
    \small
    \bibliographystyle{ieeenat_fullname}
    \bibliography{main}
}

\end{document}